\documentclass{article}
\PassOptionsToPackage{numbers, compress}{natbib}
\usepackage[dblblindworkshop, final]{neurips_2026}
\usepackage[utf8]{inputenc} 
\usepackage[T1]{fontenc}    
\usepackage{hyperref}       
\usepackage{url}            
\usepackage{booktabs}       
\usepackage{amsfonts}       
\usepackage{amsmath}        
\usepackage{nicefrac}       
\usepackage{microtype}      
\usepackage{xcolor}         
\usepackage{algorithm}
\usepackage{algpseudocode} 
\usepackage{caption}        
\usepackage{wrapfig}
\usepackage{graphicx}
\usepackage{tikz}
\usepackage{multicol}
\usepackage{algpseudocode}
\usepackage{placeins}
\usepackage{array}
\usepackage{tabularx}
\usetikzlibrary{arrows.meta,calc}

\usepackage{todonotes}
\usepackage{xcolor}

\title{Pretraining for Forecasting in Relational Databases}
\workshoptitle{Foundation Models for Temporal Systems: From Forecasting to World Modeling}

\author{%
  Ziyue Lin$^{1,2}$ \!
  Sebastian Regalado Lozano$^{1}$ \!
  \textbf{Michael Guerzhoy$^{2}$} \!
  \textbf{Ricardo Baptista$^{2}$} \!
  \textbf{George Stein$^{1}$} \\
  $^{1}$Layer 6 AI \qquad
  $^{2}$University of Toronto \\
}

\begin{document}

\maketitle
\begin{abstract}
Real-world forecasting increasingly relies on asynchronous transaction logs in relational databases rather than uniformly sampled arrays. While relational deep learning (RDL) models can forecast from these complex histories, they are typically trained separately for each specific target and root entity type. We introduce Query-Conditioned Relational pretraining (QCRP), a self-supervised framework that enables a standard RDL backbone to act as a universal forecaster across a database. QCRP generates diverse forecasting objectives using a structured temporal query grammar and interleaves them through a shared conditional predictor. This produces a robust, frozen encoder reusable for downstream tasks via lightweight predictors or a zero-shot interface. Evaluated on large-scale RelBench databases, QCRP learns frozen representations that support competitive forecasting across tasks and entity types. Furthermore, a single unified QCRP model checkpoint is competitive with isolated, entity-specific specialists. We demonstrate promising results for reusing the learned representations from standard RDL backbones for forecasting tasks in relational databases.  

\end{abstract}

\vspace{-2mm}

\section{Introduction}
\label{sec:introduction}
\vspace{-2mm}

Traditional time-series forecasting predominantly focuses on variables measured at a standard, uniform cadence. However, in practice, a vast amount of predictive modeling operates on complex temporal environments captured within relational databases. These environments are characterized by continuous, asynchronous event streams, such as transaction logs, user interactions, irregular clinical observations \citep{labach_duett_2023}, or sensor readings, rather than neatly aligned, uniformly sampled arrays. Directly forecasting future event occurrences or states from these irregular histories requires architectures capable of handling non-uniform temporal intervals and complex dependencies across multiple data modalities.


Such forecasts are typically anchored to different entity types, or ``root'' tables, such as users, products, and accounts. Although these entities share the same evolving database history, standard relational deep learning (RDL) pipelines train separate models for each target, horizon, and root entity type \citep{fey_relational_2023,robinson_relbench_2024}. Recent relational foundation models and transferable encoders seek greater reuse \citep{wang_griffin_2025,peleska_universal_2026,ranjan_rt_2026,hudovernik_kumorfm2_2026}, but often introduce specialized architectures or target cross-database generalization. We instead ask a complementary question: can a conventional relational encoder be pretrained once on the temporal history of a single database and reused across its many forecasting problems?

Self-supervised learning offers a pathway to train these conventional architectures on large-scale temporal corpora. Methods like contrastive learning and masked graph reconstruction learn structural representations, but do so without explicitly predicting future temporal outcomes \citep{peleska_ctr_2026,hou_graphmae_2022,hou_graphmae2_2023}. Alternatively, frameworks like TVE predict future relational statistics \citep{truong_pre-training_2025}, directly connecting pretraining to entity-level forecasting. Generating pseudo-tasks has also shown promise for tabular adaptation \citep{nam_stunt_2023} and tabular foundation model pretraining \citep{ma_tabdpt_2025,shaheen_generalization_2026}, while RelPrism \citep{yang_relprism_2026} broadens it to relational pretraining. These approaches successfully motivate objectives reflecting downstream predictive needs. However, TVE and RelPrism are both explicitly root-table-centric: their predictive objectives are defined for one root entity type at a time, even though different roots share the same evolving database history.
Moreover, a reusable forecaster must support a much broader space of future targets than a fixed task vector or the small set of forecasting tasks exposed by existing relational benchmarks \citep{gu_relbenchv2_2026,azevedo_taskscarcity_2026}, despite a much richer space of declaratively specifiable relational forecasting objectives \citep{kolesnichenko_rtgl_2026}. This motivates pretraining over a diverse family of temporally valid predictive objectives rather than a fixed collection of downstream labels.

We introduce Query-Conditioned Relational pretraining (QCRP), which converts structured temporal queries into self-supervised forecasting objectives for conventional RDL backbones such as GraphSAGE and RelGNN \citep{hamilton_graphsage_2018,chen_relgnn_2025}. QCRP pretrains one relational encoder on query-defined future outcomes across multiple root types, then freezes it for lightweight adaptation or zero-shot query-conditioned prediction (Fig. \ref{fig:qcrp-architecture}). Across relational forecasting benchmarks and held-out query-defined tasks, QCRP yields strong frozen transfer and positive cross-entity transfer under matched exposure, showing that conventional relational encoders can serve as reusable database-level temporal models for event-driven forecasting. The remainder of the paper is organized as follows. We describe the pretraining framework in Section~\ref{sec:method}, evaluate the frozen reuse and multi-root transfer methodology in Section~\ref{sec:experiments}, and conclude with implications and limitations for temporal relational forecasting in Section~\ref{sec:conclusion}.


\section{Multi-Root Relational Pretraining}
\label{sec:method}



\begin{figure}
    \centering
    \includegraphics[width=1\linewidth]{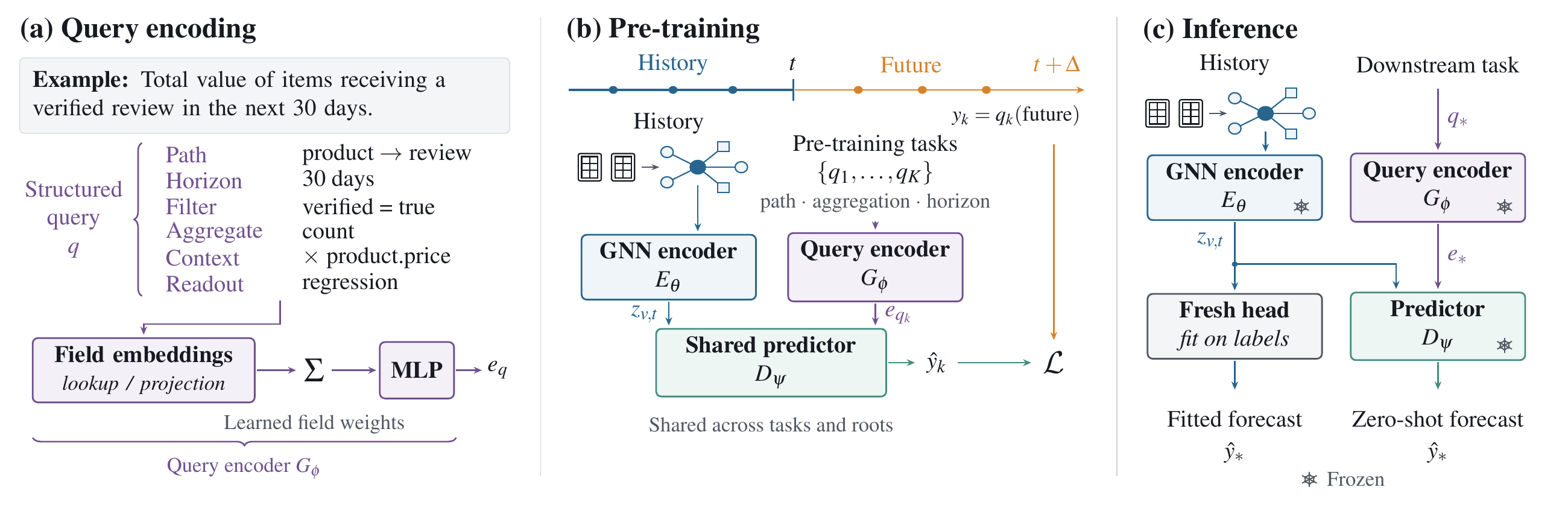}
    \vspace{-8mm}
    \caption{\textbf{QCRP overview.}
    (a) Structured temporal queries are encoded into task embeddings.
    (b) During pretraining, one historical entity representation is shared across query-defined future forecasting objectives through a conditional predictor.
    (c) At inference, the frozen relational encoder supports lightweight fitted heads, while the frozen query-conditioned predictor provides zero-shot forecasts.}
    \label{fig:qcrp-architecture}
    \vspace{-2mm}
\end{figure}

\paragraph{Relational entity-level forecasting.}
Following RDL and RelBench \citep{fey_relational_2023,robinson_relbench_2024}, let $v \in \mathcal{V}_r$ denote an entity from root table $r$, and let $t$ be a prediction cutoff. An RDL backbone $E_\theta$ maps the entity and its pre-cutoff relational history $H_{<t}$ to a $d$-dimensional representation:
$
    z_{v,t}=E_\theta(v,H_{<t}) \in \mathbb{R}^d.
$
Sampling $H_{<t}$ strictly before $t$ prevents future-event features from entering the representation. The forecasting target is defined from events in a future window $(t,t+\Delta]$. Thus, QCRP summarizes historical activity to predict diverse future outcomes (Fig.~\ref{fig:qcrp-architecture}(b)).

\vspace{-2mm}

\paragraph{Query-defined future window supervision.}
Rather than pretraining on a fixed target vector, QCRP defines scalar
forecasting objectives through structured temporal queries.
As illustrated in Fig.~\ref{fig:qcrp-architecture}(a), a query $q$ specifies the future relational quantity to predict, which is encoded as
$
    e_q = G_\eta(q) \in \mathbb{R}^p.
$
A shared conditional predictor then produces
$
    \hat y_{v,t,q}=D_\phi(z_{v,t},e_q).
$
This extends future-statistic pretraining \citep{truong_pre-training_2025} from a fixed collection of statistics to a structured family of forecasting objectives inspired by predictive query languages \citep{kocijan_pql_2026, kolesnichenko_rtgl_2026}. The relational encoder summarizes \emph{what has happened}, while the query representation specifies \emph{which future property should be predicted}. This factorization allows the expensive relational encoding $z_{v,t}$ to be computed once and reused across many forecasting queries, while the shared conditional decoder can exploit structure common to related query definitions. The complete query grammar and task inventory are given in Appendix~\ref{app:queries} and ~\ref{app:inventory}. 

\vspace{-2mm}



\paragraph{Single-root multi-task pretraining.}
QCRP processes many forecasting objectives over the same relational computation.
At each update, we sample a minibatch $\mathcal{B}$ of entity--cutoff pairs $(v,t)$ and a set
$\mathcal{S}$ of queries, encode each historical neighborhood once, and reuse the
resulting representations across all $q\in\mathcal{S}$.
For query $q$, let $m_{i,q}\in\{0,1\}$ indicate whether its target $y_{i,q}$ is
defined for anchor $i$, and let $n_q=\sum_{i\in\mathcal{B}}m_{i,q}$.
The model loss is $\mathcal{L}_q(\theta,\eta,\phi) =
    \frac{1}{n_q}\sum_{i\in\mathcal{B}}m_{i,q}\,
    \ell_q(\hat y_{i,q},y_{i,q}),$ 
using binary cross-entropy for classification objectives and Huber loss for regression.
This shared-batch design is important to save computation: the expensive temporal neighborhood sampling and relational message passing are performed only once per minibatch. Adding another objective requires only an additional lightweight decoder evaluation. Algorithm~\ref{alg:app-multitask} in Appendix~\ref{app:training} gives the complete procedure, including objective weighting and target transformations.

\vspace{-2mm}

\paragraph{Multi-root multi-task pretraining.}
A single database may support forecasts for several root entity types, such as customers and products.
We therefore extend the same model across roots using 
block-wise interleaving: training alternates between root types while sharing $E_\theta$, $G_\eta$, and $D_\phi$.
Each block trains on forecasting queries associated with one root for $T$ gradient updates, exposing the shared representation to different views of the same evolving database.
We study the resulting cross-root transfer in Section~\ref{sec:experiments}; scheduling and exposure-matching details are deferred to the Appendix ~\ref{app:scheduling}.

\vspace{-2mm}

\begin{table}[t]
\centering
\caption{\textbf{Frozen GraphSAGE representation reuse.} Mean $\pm$ SD over three seeds. LTV/sales: MAE; churn: AUC $\times100$. Pretrained encoders are frozen and receive fresh MLP probes; supervised training is end-to-end from scratch. Bold marks the best pretrained mean.}
\label{tab:fmts-transfer}
\footnotesize
\setlength{\tabcolsep}{1.5pt}
\setlength{\aboverulesep}{1.2pt}
\setlength{\belowrulesep}{1.2pt}
\renewcommand{\arraystretch}{0.94}
\begin{tabular*}{\linewidth}{@{\extracolsep{\fill}}lcccccc@{}}
\toprule
& \multicolumn{4}{c}{Rel-Amazon} & \multicolumn{2}{c}{Rel-HM} \\
\cmidrule(lr){2-5}\cmidrule(l){6-7}
Method & u-LTV$\downarrow$ & i-LTV$\downarrow$ & u-churn$\uparrow$ & i-churn$\uparrow$ & sales$\downarrow$ & u-churn$\uparrow$ \\
\midrule
Supervised & $14.707$\,{\scriptsize$\pm 0.159$} & $47.210$\,{\scriptsize$\pm 0.197$} & $69.350$\,{\scriptsize$\pm 0.030$} & $81.600$\,{\scriptsize$\pm 0.130$} & $0.041$\,{\scriptsize$\pm 0.001$} & $69.200$\,{\scriptsize$\pm 0.070$} \\
\midrule
MaskAE & $15.698$\,{\scriptsize$\pm 0.105$} & $60.135$\,{\scriptsize$\pm 0.436$} & $66.565$\,{\scriptsize$\pm 0.341$} & $64.899$\,{\scriptsize$\pm 1.715$} & $0.072$\,{\scriptsize$\pm 0.002$} & $68.161$\,{\scriptsize$\pm 0.151$} \\
CTR & $14.992$\,{\scriptsize$\pm 0.037$} & $57.527$\,{\scriptsize$\pm 0.323$} & $66.202$\,{\scriptsize$\pm 0.241$} & $76.071$\,{\scriptsize$\pm 0.422$} & $0.057$\,{\scriptsize$\pm 0.001$} & $67.112$\,{\scriptsize$\pm 0.090$} \\
TVE & $14.652$\,{\scriptsize$\pm 0.019$} & $54.544$\,{\scriptsize$\pm 0.100$} & $69.463$\,{\scriptsize$\pm 0.027$} & $81.975$\,{\scriptsize$\pm 0.033$} & $0.046$\,{\scriptsize$\pm 0.002$} & $68.476$\,{\scriptsize$\pm 0.102$} \\
TVE+CTR & $14.426$\,{\scriptsize$\pm 0.027$} & $55.096$\,{\scriptsize$\pm 0.299$} & $69.730$\,{\scriptsize$\pm 0.086$} & $81.921$\,{\scriptsize$\pm 0.069$} & $0.041$\,{\scriptsize$\pm 0.001$} & $68.724$\,{\scriptsize$\pm 0.071$} \\
QCRP\!$_{\text{root}}$ & $14.343$\,{\scriptsize$\pm 0.008$} & $\mathbf{46.203}$\,{\scriptsize$\pm 0.117$} & $\mathbf{69.976}$\,{\scriptsize$\pm 0.077$} & $82.268$\,{\scriptsize$\pm 0.061$} & $\mathbf{0.039}$\,{\scriptsize$\pm 0.000$} & $68.613$\,{\scriptsize$\pm 0.172$} \\
QCRP\!$_{\text{multi}}$ & $\mathbf{14.328}$\,{\scriptsize$\pm 0.040$} & $47.252$\,{\scriptsize$\pm 0.375$} & $69.872$\,{\scriptsize$\pm 0.031$} & $\mathbf{82.301}$\,{\scriptsize$\pm 0.055$} & $0.040$\,{\scriptsize$\pm 0.000$} & $\mathbf{68.985}$\,{\scriptsize$\pm 0.052$} \\
\bottomrule
\end{tabular*}
\par\vspace{2pt}
{\scriptsize\raggedright QCRP\!$_{\text{root}}$: separate root-specific checkpoints; QCRP\!$_{\text{multi}}$: one checkpoint per database, with matched per-root exposure. MaskAE: masked reconstruction; CTR: contrastive pretraining alone. TVE uses two-hop statistics; TVE+CTR combines TVE with contrastive learning objective ($\mathcal{L}=0.9 \mathcal{L}_{\text{TVE}} + 0.1 \mathcal{L}_{\text{CTR}}$). SDs are rounded; bold does not imply statistical significance.\par}
\end{table}

\paragraph{Frozen downstream reuse.}
After pretraining, the relational encoder is frozen and reused as shown in Fig.~\ref{fig:qcrp-architecture}(c). For a downstream task with labeled examples $\{(v_i,t_i,y_i)\}$, we compute frozen representations $z_{v_i,t_i}=E_\theta(v_i,H_{<t_i})$ and train only a two-layer MLP
$
    \hat y_i=g_\psi(z_{v_i,t_i}),
$
which outputs a regression value or classification logit. This directly tests whether QCRP has learned reusable temporal relational features rather than merely a good initialization. For query-defined tasks, the frozen query encoder and conditional predictor $D_\phi(z_{v,t},G_\eta(q))$ additionally provide a zero-shot forecasting interface.
\vspace{-2mm}
\section{Experiments and Results}
\label{sec:experiments}
\vspace{-1mm}

\paragraph{Setup.}
We follow RelBench, a widely adopted benchmark for relational learning on temporal databases \citep{robinson_relbench_2024} focusing on Rel-Amazon and Rel-HM, two large-scale datasets with forecasting tasks spanning multiple root entity types.
We evaluate six entity-level tasks using the official temporal splits, and report mean absolute error (MAE) for regression tasks, and ROC-AUC for binary classification.
Table~\ref{tab:fmts-transfer} compares GraphSAGE \citep{hamilton_graphsage_2018} pretrained with masked reconstruction (MaskAE; \citep{hou_graphmae_2022}), contrastive relational pretraining (CTR; \citep{peleska_ctr_2026}), Task Vector Estimation (TVE), and TVE combined with CTR (TVE+CTR; \citep{truong_pre-training_2025}).
Single-root QCRP\!$_{\text{root}}$ uses separately trained checkpoints for each root entity type, while multi-root QCRP\!$_{\text{multi}}$ uses one checkpoint trained jointly across all evaluated roots in a database.
For frozen adaptation, each encoder is fixed and a fresh MLP is trained on its representations using the downstream training labels; the downstream checkpoint is selected by validation performance. We also include supervised training from scratch as a baseline for comparison. GraphSAGE results average three pretraining seeds. Table~\ref{tab:fmts-published-comparison} additionally compares multi-root RelGNN reuse \citep{chen_relgnn_2025} with published results from KumoRFM-2 and RelArena \citep{hudovernik_kumorfm2_2026,hayler_relarena_2026}. 
Detailed protocols and training budgets are given in the Appendix \ref{app:setup}.

\paragraph{Future-query supervision supports representation reuse.}
With frozen GraphSAGE embeddings, QCRP\!$_{\text{root}}$ outperforms TVE on all six tasks and TVE+CTR on five; QCRP$_\text{multi}$ exceeds both on all six (Table~\ref{tab:fmts-transfer}). 
MaskAE and CTR underperform future-statistic pretraining on all six tasks, consistent with their objectives lacking explicit supervision for future-window outcomes.
The largest Amazon gain occurs on item-LTV: QCRP\!$_{\text{root}}$ achieves 46.203 MAE versus 54.544 for TVE and 55.096 for TVE+CTR.
This target combines future review count with product price, consistent with broader objective coverage helping transfer. Full fine-tuning results are in Appendix \ref{app:transfer}.

\begin{wraptable}{r}{0.60\linewidth}
\centering
\vspace{-0.5\baselineskip}
\caption{\textbf{Published baselines vs.\ multi-root QCRP.}
LTV/sales: MAE $\downarrow$; churn: AUC $\uparrow$. Top: pretrained foundation models; bottom: conventional RDL models. QCRP$_\text{multi}$ uses one pretrained RelGNN checkpoint per database with lightweight MLP probes; published methods use differing adaptation protocols: conventional RDL baselines are trained end-to-end for each task, RT-PluRel performs task-specific fine-tuning and context selection, relational tabular methods such as RDBLearn and TabPFN-Rel construct aggregate features before tabular in-context prediction.}
\label{tab:fmts-published-comparison}

\scriptsize
\setlength{\tabcolsep}{1.2pt}
\setlength{\aboverulesep}{1pt}
\setlength{\belowrulesep}{1pt}
\renewcommand{\arraystretch}{0.91}

\begin{tabular*}{\linewidth}{@{\extracolsep{\fill}}lrrrrrrr@{}}
\toprule
& \multicolumn{4}{c}{Amazon}
& \multicolumn{2}{c}{HM}
& Rank \\
\cmidrule(lr){2-5}\cmidrule(lr){6-7}
Method
& u-LTV
& i-LTV
& u-C
& i-C
& Sales
& u-C
& $\downarrow$ \\
\midrule

Griffin
& 35.59 & 53.21 & 62.30 & 69.00 & .151 & 60.20 & 10.17 \\
RDBLearn
& 14.54 & 48.56 & 67.57 & 82.07 & .064 & 68.05 & 7.58 \\
KumoRFM-2
& 13.92 & 46.99 & 69.10 & 82.17 & \textbf{.034} & 69.27 & 4.58 \\
RT-PluRel
& 13.94 & \textbf{43.03} & \textbf{71.35} & \textbf{83.27}
& .040 & 70.44 & \textbf{2.00} \\
TabPFN-Rel
& 14.36 & 46.77 & 70.86 & 82.80 & .061 & 70.52 & 4.50 \\

\midrule

LightGBM
& 16.78 & 60.57 & 52.22 & 62.54 & .076 & 55.21 & 10.67 \\
HGT
& 15.41 & 55.87 & 66.43 & 77.97 & .064 & 66.95 & 9.08 \\
GraphSAGE
& 14.31 & 50.05 & 70.42 & 82.81 & .056 & 69.88 & 5.50 \\
RelGT
& 14.27 & 48.92 & 70.39 & 82.55 & .054 & 69.27 & 5.83 \\
RelGNN$^\dagger$
& 14.23 & 48.77 & 70.99 & 82.64 & .054 & \textbf{70.93} & 3.75 \\

\midrule
\textbf{QCRP$_\text{multi}$ (RelGNN)}
& \textbf{13.83} & 46.10 & 70.87 & 82.90
& .036 & 69.96 & 2.33 \\
\bottomrule
\end{tabular*}

\vspace{1pt}
{\tiny\raggedright
 $^\dagger$Published RelGNN uses task-specific architectures/settings. 
\par}
\vspace{-0.4\baselineskip}

\end{wraptable}
\vspace{-3mm}

\paragraph{One checkpoint across root types.}
Blocked interleaving (block size $T=5000$) shares one encoder across the two root types in each database.
Comparisons match per-root exposure: a multi-root run receives a larger total budget than either specialist alone.
QCRP\!$_{\text{multi}}$ remains competitive with single-root specialist QCRP\!$_{\text{root}}$, improving three of six displayed means while losing ground on the other three (Table~\ref{tab:fmts-transfer}). These results demonstrated interleaving as a viable path for multi-root training. Scheduling sensitivity and switching dynamics are analyzed in the Appendix \ref{app:scheduling}.


With RelGNN, QCRP$_\text{multi}$ achieves the second-best average rank among the state-of-the-art
methods, behind RT-PluRel, and the best Amazon user-LTV score
(Table~\ref{tab:fmts-published-comparison}).
Together, these results show that a conventional relational encoder can be
pretrained once across entity populations and retain competitive frozen
forecasting performance without maintaining a separate encoder for each root.

\vspace{-3mm}

\paragraph{Transfer to additional forecasting objectives.}
Finally, to test whether QCRP learns reusable temporal features beyond the objectives
seen during pretraining, we evaluate 14 grammar-defined Rel-Amazon tasks whose definitions are excluded from both QCRP's and TVE's pretraining pools (Appendix~\ref{app:queries}).
An MLP fitted on held-out-task training labels achieves 81.7 mean test AUC with frozen QCRP embeddings, exceeding TVE (79.2) and matching supervised training from scratch (81.6).
The retained query encoder and decoder also permit prediction without task-specific fitting: correct query specifications achieve 68.9 mean AUC, compared with 48.8 when query fields are shuffled while inputs, labels, and weights remain fixed.
This secondary interface is useful but uneven across tasks; per-task scores and embedding geometry are provided in the Appendix \ref{app:conditioning}.
\vspace{-3mm}
\section{Discussion and Conclusion}
\label{sec:conclusion}
\vspace{-3mm}


We introduced Query-Conditioned Relational Pretraining (QCRP), a framework that translates structured future queries into self-supervised objectives. This approach enables conventional relational encoders to pretrain at the database level rather than optimizing for a single downstream target. Our results demonstrate that standard RDL backbones can achieve broad database-level reusability: a single pretrained RelGNN checkpoint successfully transfers across diverse forecasting tasks and root entity types. This complements relational foundation models, which achieve reuse via specialized large-scale architectures and pretraining regimes, by showing that standard RDL backbones can also achieve similar reuse. Query conditioning additionally enables zero-shot predictions for structured tasks. The learned task representations capture meaningful structure among forecasting objectives.

Our study also highlights limitations of current evaluation. Existing relational benchmarks provide only a small number of forecasting objectives per database, a form of \emph{task scarcity} recently identified as a bottleneck for relational transfer \citep{azevedo_taskscarcity_2026}. The six standard tasks considered here further use only one prediction horizon per database (91 days for Rel-Amazon and 7 days for Rel-HM), and several are defined by simple next-window quantities such as activity, lifetime value, or sales \citep{robinson_relbench_2024,truong_pre-training_2025}. This makes it difficult to assess generalization across target semantics and temporal scales. Richer benchmarks should provide many objectives over the same evolving system while varying target semantics, horizon length, relational depth, and temporal dynamics. Finally, QCRP does not explicitly model temporal structure. Combining database-native relational representations with these temporal signals like lagged target, trajectories, periodicity, or seasonality is a promising direction toward more general foundation models for temporal systems.

\vspace{-2mm}

\bibliographystyle{plainnat}
\bibliography{references}

@inproceedings{achille2019task2vec,
  title={Task2vec: Task embedding for meta-learning},
  author={Achille, Alessandro and Lam, Michael and Tewari, Rahul and Ravichandran, Avinash and Maji, Subhransu and Fowlkes, Charless C and Soatto, Stefano and Perona, Pietro},
  booktitle={Proceedings of the IEEE/CVF international conference on computer vision},
  pages={6430--6439},
  year={2019}
}

@inproceedings{azevedo_taskscarcity_2026,
  title={Task Scarcity and Label Leakage in Relational Transfer Learning},
  author={Azevedo, Francisco Galuppo and Loures, Clarissa Lima and Correa, Denis Oliveira},
  booktitle={3rd Workshop on Navigating and Addressing Data Problems for Foundation Models at ICLR},
  year={2026},
  url={https://openreview.net/forum?id=nI2nsMMHXp},
  doi={10.48550/arXiv.2603.29914}
}

@article{chandak_everyquery_2026,
  title={EveryQuery: Zero-Shot Clinical Prediction via Task-Conditioned Pretraining over Electronic Health Records},
  author={Chandak, Payal and Kondas, Gregory and Friedman, Liat Antwarg and Kohane, Isaac and McDermott, Matthew},
  journal={arXiv preprint arXiv:2603.07900},
  year={2026},
}

@inproceedings{chen_relgnn_2025,
    title={Rel{GNN}: Composite Message Passing for Relational Deep Learning},
    author={Tianlang Chen and Charilaos Kanatsoulis and Jure Leskovec},
    booktitle={Forty-second International Conference on Machine Learning},
    year={2025},
    url={https://openreview.net/forum?id=XXh3zmw2Uy}
}

@InProceedings{fey_relational_2023,
  title = 	 {Position: Relational Deep Learning - Graph Representation Learning on Relational Databases},
  author =       {Fey, Matthias and Hu, Weihua and Huang, Kexin and Lenssen, Jan Eric and Ranjan, Rishabh and Robinson, Joshua and Ying, Rex and You, Jiaxuan and Leskovec, Jure},
  booktitle = 	 {Proceedings of the 41st International Conference on Machine Learning},
  pages = 	 {13592--13607},
  year = 	 {2024},
  editor = 	 {Salakhutdinov, Ruslan and Kolter, Zico and Heller, Katherine and Weller, Adrian and Oliver, Nuria and Scarlett, Jonathan and Berkenkamp, Felix},
  volume = 	 {235},
  series = 	 {Proceedings of Machine Learning Research},
  month = 	 {21--27 Jul},
  publisher =    {PMLR},
  url = 	 {https://proceedings.mlr.press/v235/fey24a.html}
}

@misc{gu_relbenchv2_2026,
  title = {{RelBench} v2: A Large-Scale Benchmark and Relational Data Repository},
  author = {Gu, Justin and Ranjan, Rishabh and Kanatsoulis, Charilaos I. and Tang, Haiming and Jurkovic, Martin and Hudovernik, V. and Žnidar, M. and Chaturvedi, Pranshu and Shroff, Parth and Li, Fengyu and others},
  howpublished = {3rd Workshop on Navigating and Addressing Data Problems for Foundation Models at ICLR},
  year = {2026},
  url = {https://openreview.net/forum?id=kplDfjCbC7},
  journal = {arXiv.org},
  doi = {10.48550/arXiv.2602.12606},
}

@inproceedings{hamilton_graphsage_2018,
  title = {Inductive Representation Learning on Large Graphs},
  author = {Hamilton, William L. and Ying, Rex and Leskovec, Jure},
  booktitle = {Neural Information Processing Systems},
  volume = {30},
  year = {2017},
  url = {https://proceedings.neurips.cc/paper/2017/hash/5dd9db5e033da9c6fb5ba83c7a7ebea9-Abstract.html},
  journal = {Neural Information Processing Systems},
}

@article{hayler_relarena_2026,
  title = {Advancing Open and Reproducible Relational Learning: RelArena-$\alpha$, TabPFN-Rel and RPI},
  author={Hayler, Adrian and Fl{\"o}ge, Klemens and Arazi, Alan and Ranjan, Rishabh and Leskovec, Jure and Birkel, Felix and Roof, Brendan and Garg, Anurag and Collins, Kristina and Sidhoum, Lydia and others},
  journal={arXiv preprint arXiv:2608.16319},
  year={2026}
}

@inproceedings{hou_graphmae_2022,
  title={Graphmae: Self-supervised masked graph autoencoders},
  author={Hou, Zhenyu and Liu, Xiao and Cen, Yukuo and Dong, Yuxiao and Yang, Hongxia and Wang, Chunjie and Tang, Jie},
  booktitle={Proceedings of the 28th ACM SIGKDD conference on knowledge discovery and data mining},
  pages={594--604},
  year={2022}
}

@inproceedings{hou_graphmae2_2023,
  title={Graphmae2: A decoding-enhanced masked self-supervised graph learner},
  author={Hou, Zhenyu and He, Yufei and Cen, Yukuo and Liu, Xiao and Dong, Yuxiao and Kharlamov, Evgeny and Tang, Jie},
  booktitle={Proceedings of the ACM web conference 2023},
  pages={737--746},
  year={2023}
}

@article{hudovernik_kumorfm2_2026,
  title={KumoRFM-2: Scaling foundation models for relational learning},
  author={Hudovernik, Valter and L{\'o}pez, Federico and Kocijan, Vid and Nitta, Akihiro and Lenssen, Jan Eric and Leskovec, Jure and Fey, Matthias},
  journal={arXiv preprint arXiv:2604.12596},
  year={2026}
}

@article{kocijan_pql_2026,
  title={Predictive Query Language: A Domain-Specific Language for Predictive Modeling on Relational Databases},
  author={Kocijan, Vid and Sunil, Jinu and Lenssen, Jan Eric and Deb, Viman and Xe, Xinwei and Gomez, Federico Reyes and Fey, Matthias and Leskovec, Jure},
  journal={arXiv preprint arXiv:2602.09572},
  year={2026}
}

@article{kolesnichenko_rtgl_2026,
  title={Relational Task Generation Language: A Declarative Specification Framework for Relational Deep Learning},
  author={Kolesnichenko, Oleksii and Pele{\v{s}}ka, Jakub and others},
  journal={arXiv preprint arXiv:2609.01292},
  year={2026}
}

@article{kriegeskorte_rsa_2008,
  title={Representational similarity analysis-connecting the branches of systems neuroscience},
  author={Kriegeskorte, Nikolaus and Mur, Marieke and Bandettini, Peter A},
  journal={Frontiers in systems neuroscience},
  volume={2},
  pages={249},
  year={2008},
  publisher={Frontiers}
}

@inproceedings{
    loshchilov_adamw_2019,
    title={Decoupled Weight Decay Regularization},
    author={Ilya Loshchilov and Frank Hutter},
    booktitle={International Conference on Learning Representations},
    year={2019},
    url={https://openreview.net/forum?id=Bkg6RiCqY7},
}

@inproceedings{mahabadi_hypernet_2021,
  title={Parameter-efficient multi-task fine-tuning for transformers via shared hypernetworks},
  author={Mahabadi, Rabeeh Karimi and Ruder, Sebastian and Dehghani, Mostafa and Henderson, James},
  booktitle={Proceedings of the 59th Annual Meeting of the Association for Computational Linguistics and the 11th International Joint Conference on Natural Language Processing (Volume 1: Long Papers)},
  pages={565--576},
  year={2021}
}

@inproceedings{mayo_multitask_2023,
  title={Multitask learning via interleaving: A neural network investigation},
  author={Mayo, David and Scott, Tyler R and Ren, Mengye and Elsayed, Gamaledin and Hermann, Katherine and Jones, Matt and Mozer, Michael},
  booktitle={Proceedings of the annual meeting of the cognitive science society},
  volume={45},
  number={45},
  year={2023}
}

@inproceedings{nam_stunt_2023,
    title={{STUNT}: Few-shot Tabular Learning with Self-generated Tasks from Unlabeled Tables},
    author={Jaehyun Nam and Jihoon Tack and Kyungmin Lee and Hankook Lee and Jinwoo Shin},
    booktitle={The Eleventh International Conference on Learning Representations },
    year={2023},
    url={https://openreview.net/forum?id=_xlsjehDvlY}
}

@inproceedings{peleska_ctr_2026,
  title={Task-agnostic contrastive pretraining for relational deep learning},
  author={Pele{\v{s}}ka, Jakub and {\v{S}}{\'\i}r, Gustav},
  booktitle={Joint European Conference on Machine Learning and Knowledge Discovery in Databases},
  pages={623--638},
  year={2025},
  organization={Springer}
}

@inproceedings{peleska_universal_2026,
  title={Universal Encoders for Modular Relational Deep Learning},
  author={Pele{\v{s}}ka, Jakub},
  booktitle={Joint European Conference on Machine Learning and Knowledge Discovery in Databases},
  pages={73--90},
  year={2026},
  organization={Springer}
}

@inproceedings{ranjan_rt_2026,
 author = {Ranjan, Rishabh and Hudovernik, Valter and Znidar, Mark and Kanatsoulis, Charilaos and Upendra, Roshan and Mohammadi, Mahmoud and Meyer, Joe and Palczewski, Tom and Guestrin, Carlos and Leskovec, Jure},
 booktitle = {International Conference on Learning Representations},
 editor = {C. Vondrick and B. Hariharan and C. Raffel and L. Pinto and D. Yang and A. Faust},
 pages = {69149--69174},
 title = {Relational Transformer: Toward Zero-Shot Foundation Models for Relational Data},
 url = {https://proceedings.iclr.cc/paper_files/paper/2026/file/7107d4d2e837bde2171c6b71b5bde954-Paper-Conference.pdf},
 volume = {2026},
 year = {2026}
}

@inproceedings{robinson_relbench_2024,
 author = {Robinson, Joshua and Ranjan, Rishabh and Hu, Weihua and Huang, Kexin and Han, Jiaqi and Dobles, Alejandro and Fey, Matthias and Lenssen, Jan E. and Yuan, Yiwen and Zhang, Zecheng and He, Xinwei and Leskovec, Jure},
 booktitle = {Advances in Neural Information Processing Systems},
 doi = {10.52202/079017-0672},
 editor = {A. Globerson and L. Mackey and D. Belgrave and A. Fan and U. Paquet and J. Tomczak and C. Zhang},
 pages = {21330--21341},
 publisher = {Curran Associates, Inc.},
 title = {RelBench: A Benchmark for Deep Learning on Relational Databases},
 url = {https://proceedings.neurips.cc/paper_files/paper/2024/file/25cd345233c65fac1fec0ce61d0f7836-Paper-Datasets_and_Benchmarks_Track.pdf},
 volume = {37},
 year = {2024}
}

@inproceedings{sanh_mtprompt_2022,
    title={Multitask Prompted Training Enables Zero-Shot Task Generalization},
    author={Victor Sanh and Albert Webson and Colin Raffel and Stephen Bach and Lintang Sutawika and Zaid Alyafeai and Antoine Chaffin and Arnaud Stiegler and Arun Raja and Manan Dey and M Saiful Bari and Canwen Xu and Urmish Thakker and Shanya Sharma Sharma and Eliza Szczechla and Taewoon Kim and Gunjan Chhablani and Nihal Nayak and Debajyoti Datta and Jonathan Chang and Mike Tian-Jian Jiang and Han Wang and Matteo Manica and Sheng Shen and Zheng Xin Yong and Harshit Pandey and Rachel Bawden and Thomas Wang and Trishala Neeraj and Jos Rozen and Abheesht Sharma and Andrea Santilli and Thibault Fevry and Jason Alan Fries and Ryan Teehan and Teven Le Scao and Stella Biderman and Leo Gao and Thomas Wolf and Alexander M Rush},
    booktitle={International Conference on Learning Representations},
    year={2022},
    url={https://openreview.net/forum?id=9Vrb9D0WI4}
}

@inproceedings{standley_tasks_2020,
  title={Which tasks should be learned together in multi-task learning?},
  author={Standley, Trevor and Zamir, Amir and Chen, Dawn and Guibas, Leonidas and Malik, Jitendra and Savarese, Silvio},
  booktitle={International conference on machine learning},
  pages={9120--9132},
  year={2020},
  organization={PMLR}
}

@inproceedings{truong_pre-training_2025,
 author = {Truong, Quang and Chen, Zhikai and Ju, Mingxuan and Zhao, Tong and Shah, Neil and Tang, Jiliang},
 booktitle = {Advances in Neural Information Processing Systems},
 doi = {10.52202/085713-0277},
 editor = {D. Belgrave and C. Zhang and H. Lin and R. Pascanu and P. Koniusz and M. Ghassemi and N. Chen},
 pages = {8122--8153},
 publisher = {Curran Associates, Inc.},
 title = {A Pre-training Framework for Relational Data with Information-theoretic Principles},
 url = {https://proceedings.neurips.cc/paper_files/paper/2025/file/0c0936a505992ce0d45169cd595b9755-Paper-Conference.pdf},
 volume = {38, Main Conference},
 year = {2025}
}

@inproceedings{wang_griffin_2025,
    title={Griffin: Towards a Graph-Centric Relational Database Foundation Model},
    author={Yanbo Wang and Xiyuan Wang and Quan Gan and Minjie Wang and Qibin Yang and David Wipf and Muhan Zhang},
    booktitle={Forty-second International Conference on Machine Learning},
    year={2025},
    url={https://openreview.net/forum?id=TxeCxVb3cL}
}

@article{yang_relprism_2026,
  title={RelPrism: A Multi-Faceted Pre-training Framework with Self-Generated Tasks for Relational Databases},
  author={Yang, Jinyu and Yang, Cheng and Chen, Junze and Liu, Zedi and Zhang, Muhan and Peng, Hanyang and Shi, Chuan},
  journal={arXiv preprint arXiv:2605.23241},
  year={2026}
}

@inproceedings{you_graphctr_2021,
 author = {You, Yuning and Chen, Tianlong and Sui, Yongduo and Chen, Ting and Wang, Zhangyang and Shen, Yang},
 booktitle = {Advances in Neural Information Processing Systems},
 editor = {H. Larochelle and M. Ranzato and R. Hadsell and M.F. Balcan and H. Lin},
 pages = {5812--5823},
 publisher = {Curran Associates, Inc.},
 title = {Graph Contrastive Learning with Augmentations},
 url = {https://proceedings.neurips.cc/paper_files/paper/2020/file/3fe230348e9a12c13120749e3f9fa4cd-Paper.pdf},
 volume = {33},
 year = {2020}
}

@article{yu_pcgrad_2020,
  title={Gradient surgery for multi-task learning},
  author={Yu, Tianhe and Kumar, Saurabh and Gupta, Abhishek and Levine, Sergey and Hausman, Karol and Finn, Chelsea},
  journal={Advances in neural information processing systems},
  volume={33},
  pages={5824--5836},
  year={2020}
}

@inproceedings{zamir_taskonomy_2018,
  title={Taskonomy: Disentangling task transfer learning},
  author={Zamir, Amir R and Sax, Alexander and Shen, William and Guibas, Leonidas J and Malik, Jitendra and Savarese, Silvio},
  booktitle={Proceedings of the IEEE conference on computer vision and pattern recognition},
  pages={3712--3722},
  year={2018}
}

@inproceedings{labach_duett_2023,
  title     = {{DuETT}: Dual Event Time Transformer for
               Electronic Health Records},
  author    = {Labach, Alex and Pokhrel, Aslesha and
               Huang, Xiao Shi and Zuberi, Saba and
               Yi, Seung Eun and Volkovs, Maksims and
               Poutanen, Tomi and Krishnan, Rahul G.},
  booktitle = {Proceedings of the 8th Machine Learning
               for Healthcare Conference},
  volume    = {219},
  pages     = {403--422},
  publisher = {PMLR},
  year      = {2023},
  url       = {https://proceedings.mlr.press/v219/labach23a.html}
}

@inproceedings{ma_tabdpt_2025,
  title     = {{TabDPT}: Scaling Tabular Foundation
               Models on Real Data},
  author    = {Ma, Junwei and Thomas, Valentin and
               Hosseinzadeh, Rasa and Labach, Alex and
               Kamkari, Hamidreza and Cresswell, Jesse C. and
               Golestan, Keyvan and Yu, Guangwei and
               Caterini, Anthony L. and Volkovs, Maksims},
  booktitle = {Advances in Neural Information Processing Systems},
  volume    = {38},
  year      = {2025},
  doi       = {10.52202/085713-5748}
}

@inproceedings{shaheen_generalization_2026,
  title     = {Understanding the Surprising Generalization
               Properties of Tabular Foundation Models},
  author    = {Shaheen, Nour and Ma, Junwei and
               Labach, Alex and Hutter, Frank and
               Thomas, Valentin and Caterini, Anthony L.},
  booktitle = {Advances in Neural Information Processing Systems},
  year      = {2026},
  url       = {https://arxiv.org/abs/2608.17957}
}

\clearpage
\appendix
\section{Reproducibility}
\label{app:setup}
\subsection{Databases and prediction populations}
A relational database consists of tables of entities and events, linked by foreign keys; relational deep learning represents these tables and links as a heterogeneous graph~\citep{fey_relational_2023}. In forecasting, we choose a root entity and a prediction cutoff, and use the relational neighborhood available at that point as input. For example, to predict a customer's spending over the next month, the customer ID identifies the root node, and the model input contains the historical subgraph. This may include the customer's past purchases and linked product information available before the prediction cutoff. 

\begin{table}[ht]
\centering\small
\caption{Database contents and temporal evaluation. Counts refer to the database tables, not the number of labeled forecasting examples.}\label{tab:app-databases}
\begin{tabular}{@{}lrr@{}}\toprule
& Rel-Amazon & Rel-HM\\\midrule
Customer entities & 1,850,193 & 1,371,980\\
Product/article entities & 506,012 & 105,542\\
Review/transaction events & 20,862,040 & 15,453,651\\
Prediction horizon & 91 days & 7 days\\
First training cutoff & 2008-01-10 & 2019-09-09\\
Last training cutoff & 2015-07-02 & 2020-08-31\\
Validation cutoff & 2015-10-01 & 2020-09-07\\
Test cutoff & 2016-01-01 & 2020-09-14\\\bottomrule
\end{tabular}
\end{table}

\begin{table}[ht]\centering\small
\caption{Numbers of labeled examples in the standard forecasting tasks. Task populations differ even within one root type.}\label{tab:app-task-sizes}
\begin{tabular}{@{}llrrr@{}}\toprule Database & Task & Train & Validation & Test\\\midrule
Amazon & user-LTV & 4,708,383 & 409,792 & 351,885\\
Amazon & item-LTV & 2,707,679 & 166,978 & 178,334\\
Amazon & user-churn & 4,708,383 & 409,792 & 351,885\\
Amazon & item-churn & 2,536,014 & 177,689 & 166,842\\
HM & item-sales & 5,488,184 & 105,542 & 105,542\\
HM & user-churn & 3,832,692 & 76,556 & 74,575\\\bottomrule
\end{tabular}\end{table}

Rel-Amazon links reviews to customers and products. Rel-HM links purchases to customers and articles. We use the official RelBench temporal task tables~\citep{robinson_relbench_2024}, with the cutoffs in Table~\ref{tab:app-databases}.  The experiments reported here use one horizon per database; the grammar's ability to represent a horizon is not evidence of multi-horizon generalization.

Table~\ref{tab:app-task-sizes} reports the number of labeled examples in each task's training, validation, and test splits. Pretraining instead constructs entity--cutoff anchors directly from the database using a database-wide fixed grid spaced every 91 days on Amazon and seven days on HM, rather than reusing task-specific training tables. RelBench uses the same cutoff grid, but applies task-specific entity eligibility at each cutoff—such as requiring recent activity for churn. Consequently, its populations differ from those in the table: 6,749,915 product and 17,642,031 customer training anchors on Amazon, and 2,647,097 article and 35,309,177 customer anchors on HM. Minibatches sample these anchor pools, so update budgets count stochastic exposures rather than exhaustive passes over the populations.

\subsection{Architecture and optimization}
\label{app:training}\label{app:implementation}\label{app:evaluation_protocol}
Table~\ref{tab:app-architecture} gives the GraphSAGE~\citep{hamilton_graphsage_2018} and RelGNN~\citep{chen_relgnn_2025} backbone configurations and the shared conditioning design. Each pretrained checkpoint uses one architecture across its downstream tasks, unlike the published RelGNN reference with task-specific configurations. Table~\ref{tab:main-mlp-adaptation} provides our own supervised comparison.

\begin{table}[htbp]
\centering\small
\caption{Architecture settings for both databases. Query conditioning follows relational encoding, so sampled objectives share the same historical representations.}
\label{tab:app-architecture}
\begin{tabular}{@{}lcc@{}}
\toprule
Component & GraphSAGE & RelGNN\\
\midrule
Message-passing layers & 2 & 4\\
Entity representation width & 128 & 256\\
Aggregation & Mean & Sum\\
Attention heads & --- & 16\\
GNN dropout & 0 & 0\\
Table encoder layers / hidden width & 4 / 128 & 4 / 128\\
Table encoder output width & 128 & 256\\
Table encoder dropout & 0.2 & 0.2\\
Temporal neighbor fan-outs & $(128,64)$ & $(128,64)$\\
Neighbor selection / subgraph & Recent / bidirectional & Recent / bidirectional\\
\midrule
Query-field / embedding width & 64 / 128 & 64 / 128\\
Query encoder dropout & 0.1 & 0.1\\
Conditional MLP hidden / output width & 256 / 128 & 256 / 256\\
Conditional MLP dropout & 0.1 & 0.1\\
\bottomrule
\end{tabular}
\end{table}

The structured query encoder combines learned categorical embeddings, numerical projections, and mean-pooled schema-path embeddings using learned field weights, layer normalization, and a two-layer projection with GELU activation.

\paragraph{Shared computation and objective weighting.}
For each minibatch of $B=512$ anchors, we sample $K=16$ queries uniformly without replacement from the active root's finite training pool. If fewer than $K$ are available, all are used. Let $n_q$ be the number of defined labels for query $q$ in the minibatch, and let $\overline\ell_q$ be its mean loss over those labels. The reported configuration optimizes
\[
 \mathcal L=\frac{\sum_{q\in\mathcal S:n_q>0}w_q\overline\ell_q}
 {\sum_{q\in\mathcal S:n_q>0}w_q},\qquad
 w_q=\min\{1,\sqrt{n_q/256}\}.
\]
This reliability weight reduces the influence of objectives supported by very few examples. It is not class balancing, and the normalized weighted objective is not an unbiased estimate of an equally weighted full-pool loss. Binary objectives use binary cross-entropy with logits. Regression uses Huber loss with transition parameter one, after a $\log(1+y)$ transform for count, distinct-count, and sum targets and subsequent training-set standardization. Undefined aggregates are masked. Valid zero counts remain targets; absence and a missing measurement are not interchangeable.

Algorithm~\ref{alg:app-multitask} gives multi-task pretraining for a single root. Initialize the model and optimizer once for a standalone run. Multi-root training composes these updates by interleaving roots (Appendix~\ref{app:scheduling}); optimizer state and the learning-rate schedule persist across blocks. Future targets supply supervision only.
\begin{algorithm}[htbp]
\caption{Multi-task pretraining for one root}
\label{alg:app-multitask}
\small

\begin{algorithmic}[1]
\Require Database $\mathcal D$, root entity type $r$,
anchor pool $\mathcal A_r$, query pool $\mathcal Q_r$
\Statex \hspace{\algorithmicindent}
Relational encoder $E_\theta$, query encoder $G_\phi$,
decoder $D_\psi$, with parameters $\Theta=(\theta,\phi,\psi)$
\Statex \hspace{\algorithmicindent}
Optimizer state $s$, global step $u$,
learning-rate schedule $\{\eta_u\}$, per-query losses $\{\ell_q\}$
\Statex \hspace{\algorithmicindent}
Training-fitted target transformations,
update count $L$, batch size $B$, queries per update $K$
\Statex \hspace{\algorithmicindent} model $\Theta=(\theta,\phi,\psi)$, optimizer state $s$, global step $u$;
\Statex \hspace{\algorithmicindent} update count $L$, batch size $B$, queries per update $K$
\Procedure{MultiTaskPretrain}{$r,L,\Theta,s,u$}
\For{$j=1,\ldots,L$}
    \State Sample $B$ entity--cutoff anchors $\mathcal B$ from $\mathcal A_r$
    \State Sample $\min(K,|\mathcal Q_r|)$ queries $\mathcal S$ uniformly without replacement
    \State $Z\gets E_\theta(\{(\mathcal D_{\leq t},v):(v,t)\in\mathcal B\})$ \Comment{One encoder pass}
    \For{each $q\in\mathcal S$}
        \State $\widehat Y_q\gets D_\psi(Z,G_\phi(q))$
        \State Evaluate future targets $Y_q$ and validity mask $M_q$ on $\mathcal B$
        \State Apply training-fitted target transformations; $n_q\gets\sum M_q$
        \If{$n_q>0$}
            \State $\overline\ell_q\gets n_q^{-1}\sum_{i:M_{q,i}=1}\ell_q(\widehat Y_{q,i},Y_{q,i})$
            \State $w_q\gets\min(1,\sqrt{n_q/256})$
        \EndIf
    \EndFor
    \If{$\sum_{q:n_q>0}w_q>0$}
        \State $\mathcal L\gets\bigl(\sum_{q:n_q>0}w_q\overline\ell_q\bigr)/\bigl(\sum_{q:n_q>0}w_q\bigr)$
        \State $(\Theta,s)\gets\Call{OptimizerStep}{\Theta,s,\nabla_\Theta\mathcal L,\eta_u}$
    \EndIf
    \State $u\gets u+1$
\EndFor
\State \Return $(\Theta,s,u)$
\EndProcedure

\end{algorithmic}
\end{algorithm}

Root-specific pretraining runs for 108,000 updates, and the principal multi-root runs for 216,000 updates. GraphSAGE uses AdamW with a 5\% linear warmup followed by cosine learning-rate decay, without restarts. Peak rates are $10^{-3}$ for the Amazon specialists, $1.5\times10^{-3}$ for the Amazon multi-root run, and $5\times10^{-3}$ for HM. The reported HM RelGNN multi-root run uses $2\times10^{-4}$. Pretraining uses bfloat16 mixed precision, while downstream adaptation uses float32. Shared relational computation is performed once per minibatch; the decoder evaluates all 16 sampled objectives on those representations.

\paragraph{Evaluation regimes.}
Following the separation of representation learning and frozen-feature evaluation used in self-supervised learning, our frozen evaluation trains a fresh two-layer MLP on the entity representation, leaving the relational encoder fixed and discarding the pretraining query encoder and decoder. For the 128-dimensional GraphSAGE holdout evaluation, this predictor has 16,897 trainable parameters; the separately trained supervised model has approximately 1.36 million. Holdout probes use batch size 512, learning rate $5\times10^{-3}$, weight decay 0.01, and 20 epochs of at most 1,000 updates. Zero-shot evaluation instead retains all pretrained modules, changes only the supplied query, and performs no adaptation.

Full finetuning updates the relational encoder with a fresh linear output head. For each method and task, one rate is selected by mean validation performance over 3 pre-training seeds: $\{5\times10^{-4},10^{-3},2\times10^{-3},5\times10^{-3}\}$ on Amazon and HM. Regression checkpoints minimize validation MAE; classification checkpoints maximize validation ROC-AUC. Test values are reported only after validation selection. The standard-task tables report mean and sample standard deviation over the three seeds, except where explicitly marked as single-seed. No significance claim is inferred from mean rankings or overlapping standard deviations.

On a single NVIDIA H100 GPU, the principal 216,000-update multi-root runs take approximately 6 hours with GraphSAGE and 8 hours with RelGNN. Frozen MLP adaptation requires roughly 20 minutes per downstream task.
\FloatBarrier

\section{Task Generation and Evaluation Pools}
\label{app:queries}
\subsection{Typed query construction}

\label{sec:task_generation_grammar}

Declarative predictive queries specify targets through temporal
aggregations, filters, and comparisons \citep{kocijan_pql_2026, kolesnichenko_rtgl_2026}. We chose to construct scalar pretraining tasks
through five stages: scope, filter, aggregate, contextualize, and readout.
Each task provides both a computed target and a structured query for
conditioning the predictor.

\textbf{Scope, filter, and aggregate.}
For entity $v$ at cutoff $t$, a schema path $P$ and horizon $\Delta$
select the future relational population
$
X_{P,F}^{(t,t+\Delta]}(v)
=
\{r:\ v\xrightarrow{P}r,\ t<\tau(r)\le t+\Delta,\ F(r)\}.
$
The optional filter $F$ selects frequent categorical values through
equality predicates e.g. $\text{review.verified} =\text{true}$. We aggregate column $c$ over this population:
$
s(v,t)=A\bigl(\{c(r):r\in X_{P,F}^{(t,t+\Delta]}(v)\}\bigr).
$
Numerical columns support mean, min, max, sum, count, and standard
deviation; categorical columns support count and distinct count.

\textbf{Contextualize.}
To combine intrinsic and relational information, we optionally transform the aggregate using a root attribute available at cutoff $t$. For numerical attributes, we use $s'(v,t)=a(v,t)s(v,t)$, such as product price times future
review count. For categorical attributes, we use
$
s'(v,t)=
\operatorname{PctRank}_{u:\,a(u,t)=a(v,t)}s(u,t),
$
which measures activity relative to roots in the same category.
Without contextualization, $s'=s$.

\textbf{Readout and task selection.}
The final target is either $y_q(v,t)=s'(v,t)$ for regression or
$y_q(v,t)=\mathbf{1}[s'(v,t)\bowtie\tau]$ for binary classification,
where $\bowtie\in\{<,>,=\}$ is specified by the query.
Thresholds are zero or the 50th, 75th, and 90th percentiles of
non-null training labels. We construct a task pool under schema
and datatype constraints, excluding degenerate targets and
low-frequency predicates, and sample tasks from a balanced
distribution during pretraining without enumerating all combinations.

\paragraph{Temporal semantics of paths.}
``Direct events'' selects root-linked events in $(t,t+\Delta]$. A dimension path first selects such an event and then joins its linked entity attributes, for example customer--review--product. A same-customer or same-product path first selects a future triggering event and then joins other events sharing that entity key. The second event may not occur later than the triggering event. On HM it must additionally lie within the seven days preceding that event; the Amazon construction does not impose this lower time bound. These joins preserve multiplicities, so they are not equivalent to deduplicating all reached entities or events. The same path semantics are used across the corresponding TVE and QCRP comparisons; TVE supplies next-window aggregate supervision~\citep{truong_pre-training_2025}.

The inventory is a finite selection of generated objectives, not a fresh enumeration of all grammar combinations at each training step. Candidate predicates require sufficient training prevalence; optional categorical groups require at least 32 members. Query generation uses training-label quantiles and rejects degenerate binary targets. The selected pools contain 38 Amazon product, 40 Amazon customer, 78 HM article, and 65 HM customer objectives. HM customer supervision consists of scalar and binary objectives derived from the same two-hop aggregate families used by task-vector pretraining; the article pool additionally includes filtered and contextualized objectives. This asymmetry is an experimental design choice, not a different inference algorithm.

\subsection{Held-out task definitions and overlap}
\label{app:heldout}
The fourteen binary evaluation tasks in Table~\ref{tab:app-heldout-definitions} are excluded as complete query definitions from the 78-objective Amazon pretraining pool. All use a 91-day horizon. IDs P1--P7 and C1--C7 identify the tasks in Appendix~\ref{app:transfer}. In the table, ``verified'' denotes a verified-purchase review, ``distinct'' counts distinct observed values, and ``sd'' denotes standard deviation. Price-weighted counts multiply the \emph{root product's price} by the count; the thresholds 194.8 and 14.95 are fixed training-derived medians.
\begin{table}[htbp]
\centering\small
\caption{Held-out Rel-Amazon queries. All use a 91-day horizon. Test prevalence is computed among defined labels; undefined outcomes are excluded, not treated as negatives.}
\label{tab:app-heldout-definitions}
\renewcommand{\arraystretch}{1.06}
\begin{tabularx}{\linewidth}{@{}l>{\raggedright\arraybackslash}p{0.20\linewidth}>{\raggedright\arraybackslash}Xrr@{}}
\toprule
ID & Scope & Positive class & Test $n$ & Pos. (\%)\\
\midrule
P1 & Direct events & max(rating) $<$ 5 & 178,251 & 19.0 \\
P2 & Same product & min(rating) $>$ 3 & 178,251 & 23.8 \\
P3 & Same customer & price $\times$ count(rows) $>$ 194.8 & 178,251 & 61.6 \\
P4 & Same product & mean(rating; verified=True) $>$ 4.5 & 173,674 & 52.7 \\
P5 & Direct events & price $\times$ count(rows; verified=True) $>$ 14.95 & 178,251 & 58.4 \\
P6 & Same customer & min(rating; verified=True) $>$ 3 & 174,415 & 33.2 \\
P7 & Same product & distinct(rating; verified=True) $>$ 1 & 178,251 & 82.8 \\
C1 & Same customer & sd(rating) $>$ 1 & 371,746 & 24.2 \\
C2 & Product attributes & sd(price) $>$ 10 & 170,195 & 16.0 \\
C3 & Direct events & distinct(rating) $>$ 1 & 393,842 & 21.6 \\
C4 & Same product & sd(rating; verified=True) $>$ 1 & 382,159 & 27.9 \\
C5 & Same customer & min(rating; verified=True) $>$ 3 & 373,996 & 55.2 \\
C6 & Same customer & distinct(rating; verified=True) $>$ 2 & 393,842 & 30.0 \\
C7 & Same product & distinct(rating; verified=True) $>$ 1 & 393,842 & 93.0 \\
\bottomrule
\end{tabularx}
\end{table}

\paragraph{What is and is not held out.}
The query audit compares the complete specification, including root, path, aggregation, predicate, context, output type, and readout. Each held-out query differs from every selected training specification. Primitive values are deliberately shared, and numerical readout thresholds are allowed to be new. Five tasks introduce a new readout over a familiar computational scope, six additionally recombine one structural query axis, and three recombine two. These categories describe query construction, not a guarantee of increasing predictive difficulty.

This exclusion does \emph{not} establish independence from training targets, a concern raised in studies of task scarcity and label overlap in relational transfer~\citep{azevedo_taskscarcity_2026}. For example, P1 tests whether maximum rating is below five, while maximum rating itself is a regression objective. A model capable of estimating that aggregate may support the unseen thresholded target. Similar relations exist between distinct-count objectives and diversity tests. The valid claim is transfer to withheld query definitions, including new readouts and compositions; it is not transfer to functions with no recoverable relation to the training task pool. Nor do we claim that all exact or approximate statistical proxies have been removed. The standard benchmark tasks are not subjected to the same exclusion: direct inactivity and count statistics remain in pretraining.
\FloatBarrier

\section{Additional Results}
\label{app:transfer}
\subsection{Complete standard-task comparisons}
The baselines include task-vector estimation (TVE)~\citep{truong_pre-training_2025} and adaptations of masked graph reconstruction~\citep{hou_graphmae_2022} and graph contrastive learning~\citep{you_graphctr_2021}. Table~\ref{tab:main-mlp-adaptation} reports the frozen-transfer results for both backbones, including random initialization and our own supervised references. Table~\ref{tab:full-finetune} separates end-to-end adaptation from frozen reuse. The latter changes both the trainable encoder parameters and the predictor class: it uses a linear head rather than the frozen protocol's MLP. It therefore measures an adaptation regime, not an isolated effect of unfreezing one module.

Task grouping can produce both cooperation and competition~\citep{standley_tasks_2020}. In our experiments, multi-root reuse trades some specialist performance for one shared checkpoint; differences vary by task, with HM churn remaining the main weakness relative to supervised training.

\begin{table}[!ht]
\centering
\caption{Frozen-encoder MLP probes. Supervised training from scratch (Sup.) is an end-to-end reference; all other rows fit a fresh MLP on frozen encoder embeddings. AUC is multiplied by 100. Bold marks the best mean among pretrained rows within each backbone. }
\label{tab:main-mlp-adaptation}
\footnotesize
\setlength{\tabcolsep}{1pt}
\renewcommand{\arraystretch}{1.12}
\begin{tabular*}{\linewidth}{@{\extracolsep{\fill}}lcccccc@{}}
\toprule
& \multicolumn{4}{c}{Rel-Amazon} & \multicolumn{2}{c}{Rel-HM} \\
\cmidrule(lr){2-5}\cmidrule(l){6-7}
Initialization & user-LTV & item-LTV & user-churn & item-churn & item-sales & user-churn \\
& MAE $\downarrow$ & MAE $\downarrow$ & AUC $\uparrow$ & AUC $\uparrow$ & MAE $\downarrow$ & AUC $\uparrow$ \\
\midrule
\multicolumn{7}{l}{\textit{GraphSAGE}} \\
Random initialization & $16.444$\,{\scriptsize$\pm 0.029$} & $60.431$\,{\scriptsize$\pm 0.257$} & $65.772$\,{\scriptsize$\pm 0.062$} & $74.604$\,{\scriptsize$\pm 0.474$} & $0.051$\,{\scriptsize$\pm 0.000$} & $66.616$\,{\scriptsize$\pm 0.487$} \\
Supervised & $14.707$\,{\scriptsize$\pm 0.159$} & $47.210$\,{\scriptsize$\pm 0.197$} & $69.350$\,{\scriptsize$\pm 0.030$} & $81.600$\,{\scriptsize$\pm 0.130$} & $0.041$\,{\scriptsize$\pm 0.001$} & $69.200$\,{\scriptsize$\pm 0.070$} \\
Masked reconstruction & $15.698$\,{\scriptsize$\pm 0.105$} & $60.135$\,{\scriptsize$\pm 0.436$} & $66.565$\,{\scriptsize$\pm 0.341$} & $64.899$\,{\scriptsize$\pm 1.715$} & $0.072$\,{\scriptsize$\pm 0.002$} & $68.161$\,{\scriptsize$\pm 0.151$} \\
CTR & $14.992$\,{\scriptsize$\pm 0.037$} & $57.527$\,{\scriptsize$\pm 0.323$} & $66.202$\,{\scriptsize$\pm 0.241$} & $76.071$\,{\scriptsize$\pm 0.422$} & $0.057$\,{\scriptsize$\pm 0.001$} & $67.112$\,{\scriptsize$\pm 0.090$} \\
TVE-2 & $14.652$\,{\scriptsize$\pm 0.019$} & $54.544$\,{\scriptsize$\pm 0.100$} & $69.463$\,{\scriptsize$\pm 0.027$} & $81.975$\,{\scriptsize$\pm 0.033$} & $0.046$\,{\scriptsize$\pm 0.002$} & $68.476$\,{\scriptsize$\pm 0.102$} \\
TVE-2+CTR & $14.426$\,{\scriptsize$\pm 0.027$} & $55.096$\,{\scriptsize$\pm 0.299$} & $69.730$\,{\scriptsize$\pm 0.086$} & $81.921$\,{\scriptsize$\pm 0.069$} & $0.041$\,{\scriptsize$\pm 0.001$} & $68.724$\,{\scriptsize$\pm 0.071$} \\
\textbf{QCRP$_\text{multi}$} & $\mathbf{14.328}$\,{\scriptsize$\pm 0.040$} & $47.252$\,{\scriptsize$\pm 0.375$} & $69.872$\,{\scriptsize$\pm 0.031$} & $\mathbf{82.301}$\,{\scriptsize$\pm 0.055$} & $0.040$\,{\scriptsize$\pm 0.000$} & $\mathbf{68.985}$\,{\scriptsize$\pm 0.052$} \\
\textbf{QCRP$_\text{root}$} & $14.343$\,{\scriptsize$\pm 0.008$} & $\mathbf{46.203}$\,{\scriptsize$\pm 0.117$} & $\mathbf{69.976}$\,{\scriptsize$\pm 0.077$} & $82.268$\,{\scriptsize$\pm 0.061$} & $\mathbf{0.039}$\,{\scriptsize$\pm 0.000$} & $68.613$\,{\scriptsize$\pm 0.172$} \\
\midrule
\multicolumn{7}{l}{\textit{RelGNN}} \\
Random initialization & $16.414$\,{\scriptsize$\pm 0.018$} & $60.325$\,{\scriptsize$\pm 0.351$} & $64.721$\,{\scriptsize$\pm 0.228$} & $73.412$\,{\scriptsize$\pm 0.757$} & $0.050$\,{\scriptsize$\pm 0.001$} & $66.527$\,{\scriptsize$\pm 0.381$} \\
Supervised & $14.123$\,{\scriptsize$\pm 0.012$} & $47.897$\,{\scriptsize$\pm 0.183$} & $70.782$\,{\scriptsize$\pm 0.040$} & $82.627$\,{\scriptsize$\pm 0.038$} & $0.049$\,{\scriptsize$\pm 0.002$} & $70.634$\,{\scriptsize$\pm 0.259$} \\
TVE-2 & $14.040$\,{\scriptsize$\pm 0.016$} & $52.498$\,{\scriptsize$\pm 0.388$} & $70.861$\,{\scriptsize$\pm 0.042$} & $\mathbf{83.030}$\,{\scriptsize$\pm 0.036$} & $0.040$\,{\scriptsize$\pm 0.001$} & $69.441$\,{\scriptsize$\pm 0.186$} \\
TVE-2+CTR & $13.950$\,{\scriptsize$\pm 0.028$} & $52.442$\,{\scriptsize$\pm 0.352$} & $70.957$\,{\scriptsize$\pm 0.008$} & $82.952$\,{\scriptsize$\pm 0.014$} & $0.038$\,{\scriptsize$\pm 0.000$} & $69.839$\,{\scriptsize$\pm 0.094$} \\
\textbf{QCRP$_\text{multi}$} & $13.834$\,{\scriptsize$\pm 0.024$} & $46.103$\,{\scriptsize$\pm 1.690$} & $70.874$\,{\scriptsize$\pm 0.027$} & $82.902$\,{\scriptsize$\pm 0.035$} & $\mathbf{0.036}$\,{\scriptsize$\pm 0.001$} & $\mathbf{69.955}$\,{\scriptsize$\pm 0.156$} \\
\textbf{QCRP$_\text{root}$} & $\mathbf{13.804}$\,{\scriptsize$\pm 0.010$} & $\mathbf{45.058}$\,{\scriptsize$\pm 0.096$} & $\mathbf{70.979}$\,{\scriptsize$\pm 0.075$} & $82.995$\,{\scriptsize$\pm 0.039$} & $0.036 \pm 0.001$ & $69.749 \pm 0.098$ \\
\bottomrule
\end{tabular*}
\par\vspace{2pt}
{\footnotesize\raggedright Entries are mean $\pm$ standard deviation over three pretraining seeds. QCRP$_\text{multi}$ interleaves the two root pools ($T=5000$ updates, 216 epochs), with nominally matched per-root budgets; Appendix~\ref{app:scheduling} gives the exact update counts.\par}
\end{table}

\begin{table}[t]
\centering
\caption{Full finetuning with GraphSAGE. Each method transfers only its relational encoder, discards pretraining-specific modules, and trains the encoder with a fresh linear head. AUC is multiplied by 100.}
\label{tab:full-finetune}
\footnotesize
\setlength{\tabcolsep}{1pt}
\renewcommand{\arraystretch}{1.12}
\begin{tabular*}{\linewidth}{@{\extracolsep{\fill}}lcccccc@{}}
\toprule
& \multicolumn{4}{c}{Rel-Amazon} & \multicolumn{2}{c}{Rel-HM} \\
\cmidrule(lr){2-5}\cmidrule(l){6-7}
Initialization & user-LTV & item-LTV & user-churn & item-churn & item-sales & user-churn \\
& MAE $\downarrow$ & MAE $\downarrow$ & AUC $\uparrow$ & AUC $\uparrow$ & MAE $\downarrow$ & AUC $\uparrow$ \\
\midrule
Masked reconstruction & $15.537$\,{\scriptsize$\pm 1.110$} & $46.833$\,{\scriptsize$\pm 0.107$} & $68.820$\,{\scriptsize$\pm 0.232$} & $81.621$\,{\scriptsize$\pm 0.142$} & $0.040$\,{\scriptsize$\pm 0.001$} & $69.764$\,{\scriptsize$\pm 0.090$} \\
CTR & $14.418$\,{\scriptsize$\pm 0.051$} & $46.338$\,{\scriptsize$\pm 0.686$} & $69.347$\,{\scriptsize$\pm 0.051$} & $81.494$\,{\scriptsize$\pm 0.026$} & $0.041$\,{\scriptsize$\pm 0.001$} & $69.486$\,{\scriptsize$\pm 0.109$} \\
TVE-2 & $14.419$\,{\scriptsize$\pm 0.064$} & $46.382$\,{\scriptsize$\pm 0.515$} & $69.471$\,{\scriptsize$\pm 0.170$} & $81.799$\,{\scriptsize$\pm 0.064$} & $0.042$\,{\scriptsize$\pm 0.001$} & $69.531$\,{\scriptsize$\pm 0.127$} \\
TVE-2+CTR & $14.353$\,{\scriptsize$\pm 0.017$} & $47.098$\,{\scriptsize$\pm 0.425$} & $\mathbf{69.801}$\,{\scriptsize$\pm 0.084$} & $81.769$\,{\scriptsize$\pm 0.023$} & $\mathbf{0.038}$\,{\scriptsize$\pm 0.001$} & $\mathbf{69.837}$\,{\scriptsize$\pm 0.018$} \\
\textbf{QCRP$_\text{multi}$} & $14.511$\,{\scriptsize$\pm 0.092$} & $46.339$\,{\scriptsize$\pm 0.355$} & $69.648$\,{\scriptsize$\pm 0.154$} & $\mathbf{81.909}$\,{\scriptsize$\pm 0.032$} & $0.040$\,{\scriptsize$\pm 0.001$} & $69.809$\,{\scriptsize$\pm 0.231$} \\
\textbf{QCRP$_\text{root}$} & $\mathbf{14.333}$\,{\scriptsize$\pm 0.004$} & $\mathbf{46.228}$\,{\scriptsize$\pm 0.544$} & $69.741$\,{\scriptsize$\pm 0.127$} & $81.636$\,{\scriptsize$\pm 0.261$} & $0.038$\,{\scriptsize$\pm 0.001$} & $69.425$\,{\scriptsize$\pm 0.015$} \\
\bottomrule
\end{tabular*}
\par\vspace{2pt}
{\footnotesize\raggedright Mean $\pm$ sample standard deviation across seeds 43/44/45. One learning rate per method and task is selected by mean validation performance. The grids are $\{5\!\times\!10^{-4},10^{-3},2\!\times\!10^{-3},5\!\times\!10^{-3}\}$. All methods except QCRPmulti use root-specific checkpoints. Bold denotes a mean ranking, not statistical significance: on HM item-sales the bold TVE-2+CTR mean ($0.03786$) and QCRP ($0.03820$) differ by $0.5$ combined standard errors and print identically at this precision. QCRP$_\text{multi}$ selects lr $2\times10^{-3}$ on HM user-churn and $10^{-3}$ on HM item-sales.\par}
\end{table}

\paragraph{A difficult transfer case: HM customer churn.}
HM churn evaluates recently active customers, whereas pretraining uses a broader population. A recency-only validation diagnostic falls from 71.536 AUC on broad anchors with history to 55.216 on the recent-active cohort. Only 1.443\% of positive--negative pairs in the broad-population AUC have both endpoints in that cohort, versus 10.113\% and 19.753\% for Amazon customers and products. Broad-population discrimination can therefore obscure weak ranking within the downstream cohort. This descriptive mismatch does not isolate causality; frozen HM churn remains below supervised training, while full finetuning narrows the gap.

\subsection{Held-out objectives}
\begin{table}[htbp]\centering\small
\caption{Held-out-task test ROC-AUC ($\times100$), one seed. Random initialization, TVE, and QCRP fit fresh MLPs on frozen entity representations; supervised training updates the full model. Query IDs are defined in Table~\ref{tab:app-heldout-definitions}.}\label{tab:app-heldout-scores}
\begin{tabular}{@{}lrrrr@{}}\toprule
ID & Random & TVE & Supervised & QCRP\\\midrule
P1 & 63.88 & 72.00 & 70.82 & 72.45\\
P2 & 81.75 & 96.14 & 97.38 & 97.45\\
P3 & 68.34 & 72.96 & 77.15 & 77.46\\
P4 & 66.77 & 91.29 & 96.96 & 96.39\\
P5 & 68.14 & 73.88 & 87.58 & 85.60\\
P6 & 65.56 & 69.88 & 69.60 & 70.28\\
P7 & 89.06 & 94.71 & 97.14 & 97.24\\
\midrule
C1 & 61.98 & 91.94 & 93.68 & 94.08\\
C2 & 58.70 & 63.99 & 64.81 & 64.56\\
C3 & 61.13 & 68.08 & 67.67 & 68.77\\
C4 & 52.19 & 58.24 & 59.78 & 59.50\\
C5 & 70.16 & 91.34 & 93.84 & 93.36\\
C6 & 79.76 & 92.99 & 95.70 & 95.20\\
C7 & 64.27 & 71.59 & 70.26 & 71.73\\
\midrule
Mean & 67.98 & 79.22 & 81.60 & 81.72\\
\bottomrule\end{tabular}\end{table}

On the fourteen held-out tasks defined in Appendix~\ref{app:heldout}, frozen QCRP reaches 81.72 mean ROC-AUC, compared with 79.22 for TVE, 67.98 for random initialization, and 81.60 for supervised training (Table~\ref{tab:app-heldout-scores}). QCRP improves over TVE on all fourteen tasks, supporting representation reuse across withheld query definitions while achieving an average score comparable to task-specific supervised training.

\FloatBarrier

\section{Multi-Root Training Dynamics}
\label{app:scheduling}
\subsection{Interleaving algorithm and exposure}
Following the interleaved multi-task learning setting of \citet{mayo_multitask_2023}, Algorithm~\ref{alg:app-interleaving} applies the multi-task updates of Algorithm~\ref{alg:app-multitask} to one root at a time, retaining shared parameters and optimizer state across blocks. Main runs use block size $T=5000$, batch size 512, and 16 queries per update.
\begin{algorithm}[htbp]
\caption{Multi-root pretraining by interleaving}
\label{alg:app-interleaving}
\small
\begin{algorithmic}[1]
\Require Ordered roots $(r_0,\ldots,r_{R-1})$, update budget $U$, block size $T$
\State Initialize shared model $\Theta$, optimizer state $s$, and global step $u\gets0$
\While{$u<U$}
    \State $r\gets r_{\lfloor u/T\rfloor\bmod R}$; \quad $L\gets\min(T,U-u)$
    \State $(\Theta,s,u)\gets\Call{MultiTaskPretrain}{r,L,\Theta,s,u}$ \Comment{Algorithm~\ref{alg:app-multitask}}
\EndWhile
\State \Return $\Theta$ \Comment{One checkpoint for all roots}
\end{algorithmic}
\end{algorithm}

For $U$ updates and two alternating roots, the number of updates $U_i$ root $i\in\{0,1\}$ receives is:
\[
 U_i=T\sum_{b=0}^{\lfloor U/T\rfloor-1}\mathbf1[b\bmod2=i]
 +(U\bmod T)\mathbf1[\lfloor U/T\rfloor\bmod2=i].
\]
The main $U=216{,}000$ runs allocate 110,000 and 106,000 updates to the first and second roots, respectively, versus 108,000 per specialist. Aggregate exposure is matched; each root differs from its specialist by 1.85\%. Amazon starts with products, HM with customers.

\subsection{Inactive-root degradation and recovery}
\begin{figure}[htbp]\centering
\includegraphics[width=\linewidth]{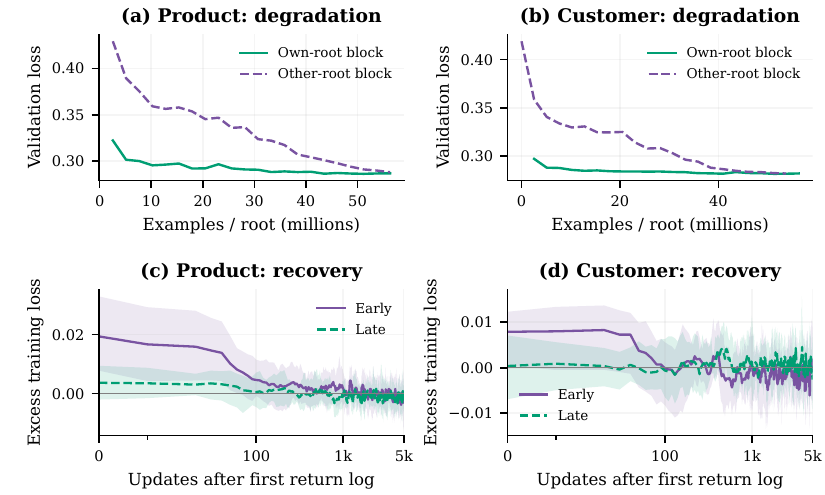}
\caption{Rel-Amazon GraphSAGE interleaving $T=5000$. Top: root-specific validation immediately after its own block and after the other root's block, at matched root exposure. Bottom: excess training loss after returning, relative to the last ten logs of the preceding active block. Early/late curves average the first/last thirds of complete resumed blocks; shading spans their 10th--90th percentiles after nine-log smoothing. Offset zero is the first return log.}\label{fig:app-amazon-scheduling}
\end{figure}

\begin{figure}[htbp]\centering
\includegraphics[width=\linewidth]{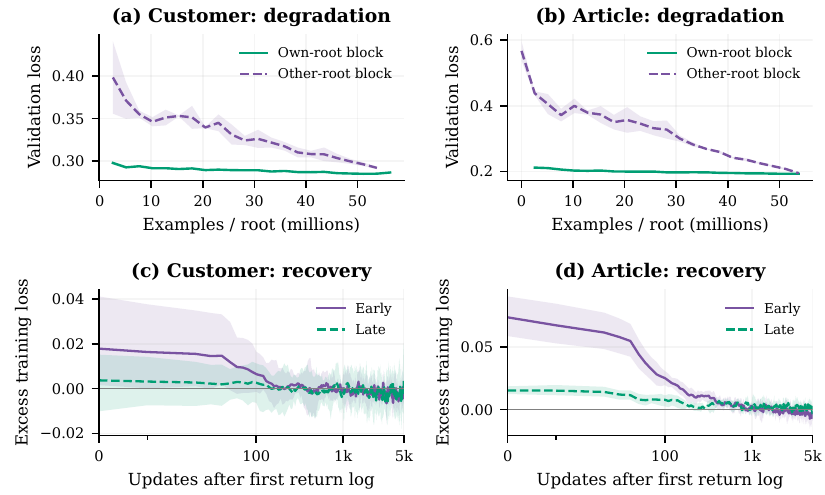}
\caption{Rel-HM GraphSAGE interleaving $T=5000$. Top: mean validation losses over two seeds, with shading spanning seeds. Bottom: recovery with the same block selection, smoothing, and percentile intervals as Amazon. Loss scales differ across roots and databases.}\label{fig:app-hm-scheduling}
\end{figure}

Figures~\ref{fig:app-amazon-scheduling} and~\ref{fig:app-hm-scheduling} show the training dynamic of multiroot pretraining by interleaving on Rel-Amazon and Rel-HM databases. Validation after the other root's block measures temporary degradation; training loss after returning measures recovery. We observe a similar phenomenon of forgetting and relearning as described in \citet{mayo_multitask_2023}. Both databases show diminishing degradation and smaller recovery transients as training progresses.

\FloatBarrier
\subsection{Joint training, block size, and gradient alignment}
Gradient interference is a known challenge in multi-task optimization~\citep{yu_pcgrad_2020}; we examine it alongside update scheduling. The Amazon comparison uses 220,000 updates for both schedules: joint training samples 256 anchors and 16 queries per root, while interleaving samples 512 anchors and 16 queries from one root. Both expose each root to 56.32 million anchors. Interleaving reaches lower final validation loss for products (0.2881 versus 0.3060) and customers (0.2822 versus 0.3011), although products are inactive at that final evaluation.

\begin{figure}[htbp]\centering
\includegraphics[width=\linewidth]{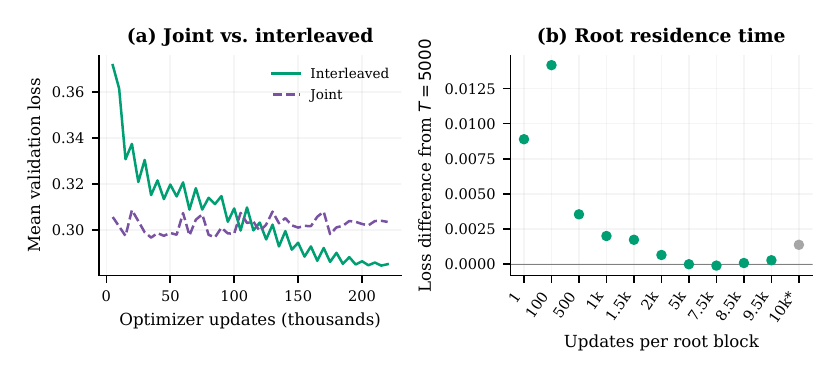}
\caption{Rel-Amazon scheduling. (a) Mean loss over 78 validation objectives; joint training sums the two root losses. Interleaved validation occurs at block boundaries and averages one freshly trained and one inactive root. (b) Residence-time sweep, 216,000 updates, relative to $T=5000$. Points are distinct settings, not a continuous fit.}
\label{fig:app-schedule-comparison}
\end{figure}

Moderate-to-long blocks form a broad favorable range (Figure~\ref{fig:app-schedule-comparison}). The comparison changes per-root batch size and loss scaling as well as update order, so it does not isolate a scheduling mechanism. Rel-HM has no corresponding joint-training run.

\begin{figure}[htbp]\centering
\includegraphics[width=\linewidth]{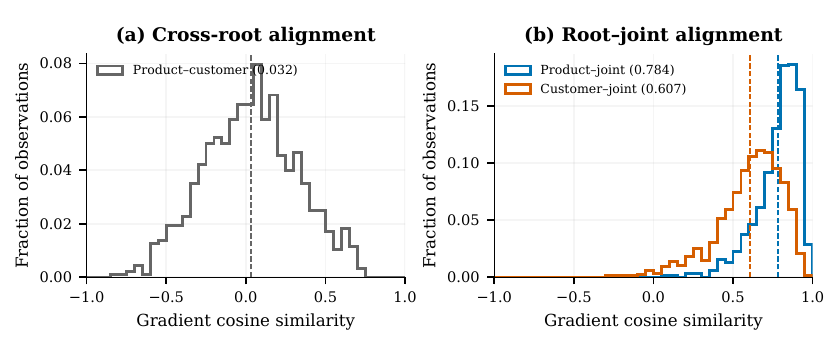}
\caption{Gradient alignment in a separate Amazon joint-training diagnostic: 880 observations, one every 250 updates, restricted to parameters used by both root losses. Histograms share cosine bins; dashed lines and legend values give means. (a) Cross-root cosine. (b) Each root gradient's cosine with their sum. The mean product/customer norm ratio is 1.406. Cosines diagnose gradient relationships~\citep{yu_pcgrad_2020}; they are not AdamW parameter updates~\citep{loshchilov_adamw_2019}.}
\label{fig:app-gradient-alignment}
\end{figure}

Root gradients are nearly orthogonal on average, while both are positively aligned with their sum.

\FloatBarrier
\subsection{Parameter-efficient adaptation and checkpoint reuse}
\label{app:parameter_efficiency}
\paragraph{Task-specific adaptation capacity}
Frozen adaptation fits a fresh predictor $h_{\omega_j}(z_{v,t})$ for task $j$ while keeping the historical encoder $E_\theta$ fixed. The query encoder and pretraining decoder are not needed. This shares the goal of limiting task-specific capacity with parameter-efficient multi-task adaptation~\citep{mahabadi_hypernet_2021}, but trains only the output predictor. Table~\ref{tab:app-adaptation-parameters} contrasts the small task-specific optimization problem with the full relational model.

\begin{table}[htbp]
\centering\small
\caption{Parameters per downstream model. Full RelGNN counts include the frozen encoder and one MLP, excluding buffers and pretraining-only modules. GraphSAGE counts refer to the fourteen-task holdout evaluation; its supervised count is rounded.}
\label{tab:app-adaptation-parameters}
\begin{tabular}{@{}llrr@{}}
\toprule
Backbone / evaluation & Adaptation & Full model & Trainable\\
\midrule
GraphSAGE, holdout & Supervised & $\approx 1.36$M & $\approx 1.36$M\\
GraphSAGE, holdout & Frozen + MLP & --- & 16,897\\
RelGNN, Amazon & Frozen + MLP & 44,396,033 & 66,561\\
RelGNN, HM & Frozen + MLP & 45,282,817 & 66,561\\
\bottomrule
\end{tabular}
\end{table}

The GraphSAGE probe trains approximately 1.2\% as many parameters as its supervised reference, while multi-root QCRP reaches 81.71 mean holdout ROC-AUC versus 81.60 for supervised training (Table~\ref{tab:app-heldout-scores}). RelGNN adapts approximately 0.15\% of its downstream model parameters on either database. Its multi-root frozen model exceeds our supervised reference on five of six standard-task means, with HM user-churn the exception (Table~\ref{tab:main-mlp-adaptation}).

\paragraph{Sharing across tasks and root types}
For each backbone and seed, one multi-root checkpoint serves customer and product tasks on Amazon, or customer and article tasks on HM. Each task adds only a predictor (Table~\ref{tab:app-checkpoint-reuse}).

\begin{table}[htbp]
\centering\small
\caption{Distinct encoder weight sets for the standard-task suite, per backbone and seed. Amazon uses four task-specific predictors; HM uses two.}
\label{tab:app-checkpoint-reuse}
\begin{tabular}{@{}lrr@{}}
\toprule
Encoder deployment & Rel-Amazon & Rel-HM\\
\midrule
Separate supervised or fully finetuned models & 4 & 2\\
Frozen root-specific models & 2 & 2\\
Frozen multi-root model & 1 & 1\\
\bottomrule
\end{tabular}
\end{table}

With encoder size $P_E$ and predictor sizes $P_{H_j}$, frozen multi-root reuse retains $P_E+\sum_{j=1}^{T}P_{H_j}$ distinct parameters for $T$ tasks. Separate task-adapted models retain a separate encoder for every task. Sharing therefore limits the additional model parameters needed for new targets, while allowing task-dependent tradeoffs relative to root-specific models (Appendix~\ref{app:transfer}).

\paragraph{Offline representation reuse}
A fixed encoder permits representations to be cached by entity and cutoff, after which task-specific predictors can be trained and evaluated without repeated relational encoding. Tasks with identical historical inputs can share the cache. Reuse requires a fixed checkpoint, preprocessing, and neighborhood construction; a new cutoff generally requires a new representation. With stochastic neighborhood sampling, caching fixes the sampled realization.

\FloatBarrier

\section{Query Conditioning}
\label{app:conditioning}
\subsection{Query embedding geometry}
Taskonomy and Task2Vec study relationships among learning tasks~\citep{zamir_taskonomy_2018,achille2019task2vec}; here we ask whether query-embedding geometry reflects empirical label relationships. For each root, labels are evaluated on a common sample of 200,000 entity--cutoff pairs. Let $r_{ij}$ be Pearson label correlation over jointly valid anchors and $c_{ij}$ query-embedding cosine. We report $\rho_{\rm signed}=\operatorname{Spearman}(r_{ij},c_{ij})$ and $\rho_{\rm absolute}=\operatorname{Spearman}(|r_{ij}|,c_{ij})$: agreement in output direction and strength of label association, respectively. Cross-root pairs and undefined correlations are excluded. Comparing pairwise similarity structures follows the representational-similarity analysis approach in~\citep{kriegeskorte_rsa_2008}.

\begin{figure}[htbp]\centering
\includegraphics[width=\linewidth]{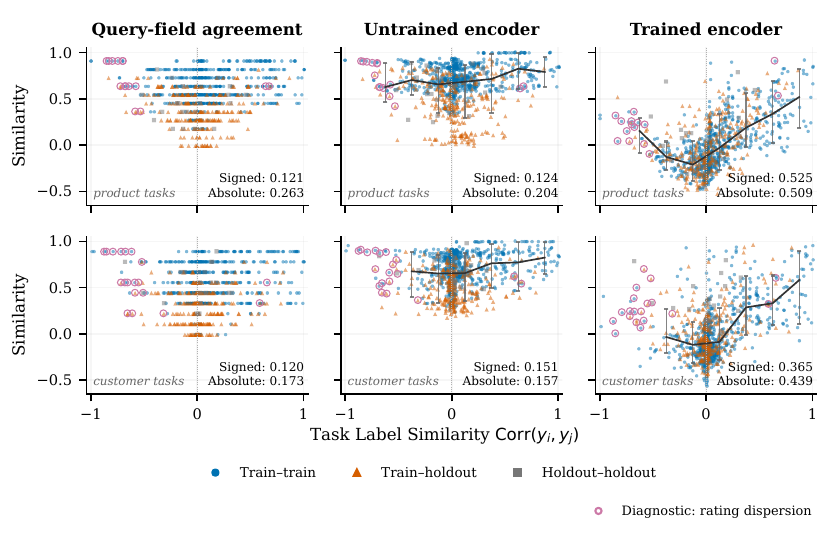}
\caption{Query semantics, embedding geometry, and task-label similarity. Columns compare raw query-field agreement, an untrained structured encoder (seed 100), and the trained encoder; rows show product and customer tasks. Field agreement is the fraction of matching variable fields; encoder similarity is cosine. Lines show medians with 10th--90th percentile spread in eight shared bins (at least 25 pairs per displayed bin); no trend is fitted to field agreement. Signed and absolute Spearman alignment appear at bottom right. Pink rings identify the post-hoc rating-dispersion diagnostic.}\label{fig:app-query-geometry}
\end{figure}
\begin{table}[ht]\centering\small
\caption{Query-geometry alignment by evaluation subset. ``Initial'' is the mean over five fresh encoders with the same architecture and vocabulary (seeds 100--104); these are not saved original initializations. Each cell lists signed / absolute-label Spearman alignment.}\label{tab:app-geometry}
\begin{tabular}{@{}llrrr@{}}\toprule Root & Pair subset & Query fields & Initial & Trained\\\midrule
Product & All within-root & .121 / .263 & .091 / .195 & .525 / .509\\
Product & Train--train & .125 / .294 & .107 / .222 & .579 / .554\\
Product & Train--holdout & -.074 / .439 & -.159 / .303 & .443 / .386\\
Customer & All within-root & .120 / .173 & .106 / .140 & .365 / .439\\
Customer & Train--train & .108 / .228 & .103 / .192 & .454 / .441\\
Customer & Train--holdout & .049 / .220 & .007 / .151 & .146 / .394\\\bottomrule
\end{tabular}\end{table}

Training strengthens pooled alignment beyond both query-field overlap and the untrained architecture. Table~\ref{tab:app-geometry} averages five fresh untrained encoders (seeds 100--104); Figure~\ref{fig:app-query-geometry} shows seed 100. The negative-correlation branch has a localized interpretation: 14 of 15 regression pairs with $r_{ij}<-0.3$ and positive embedding cosine compare rating dispersion with mean or minimum rating on the same path. Related computations can have oppositely correlated outputs without requiring opposite embeddings.

Task pairs share endpoints. Following matrix-label randomization in representational-similarity analysis~\citep{kriegeskorte_rsa_2008}, an exploratory test permutes task identities within the training and held-out groups, jointly for trained and initial matrices (1,000 permutations). On train--holdout pairs, the signed alignment difference between trained and initial encoders has two-sided $p=0.001$ for products and $0.175$ for customers; absolute alignment gives $p=0.455$ and $0.029$. These unadjusted tests support generalization that depends on the root and alignment statistic.

\subsection{Zero-shot decoding of seen queries}
Here zero-shot denotes reuse of the pretrained decoder without downstream adaptation; the query definitions themselves were seen during pretraining. Table~\ref{tab:app-seen-query-decoding} summarizes direct decoding on temporal validation anchors for the 78-query specialist experiment, separately from unseen-query evaluation below.
\begin{table}[htbp]\centering\small
\caption{Direct decoding of the 78 seen pretraining queries on temporal validation anchors, GraphSAGE specialists, one seed. Last recorded validation at 96,000 updates (not the final checkpoint). Metrics are unweighted task means; regression $R^2$ uses the transformed pretraining targets.}
\label{tab:app-seen-query-decoding}
\begin{tabular}{@{}lrrrr@{}}\toprule
Root & Binary queries & AUC $\uparrow$ & Regression queries & $R^2\uparrow$\\\midrule
Product & 12 & 82.93 & 26 & 0.368\\
Customer & 12 & 74.74 & 28 & 0.205\\
\bottomrule\end{tabular}\end{table}

\paragraph{Standard forecasting targets.}
Table~\ref{tab:app-standard-zeroshot} gives the recorded evaluation of pretraining queries on the four standard Amazon test tasks. These earlier models used a 74-query pool and are not the checkpoints underlying the main frozen-probe comparison. Regression predictions invert the pretraining-fitted normalization and log transform. For churn, the supplied query predicts future activity, so $1-p(\mathrm{activity})$ gives the churn score by definition; this is not a sign chosen from test performance. Neither downstream-label calibration nor a fitted predictor is used.
\begin{table}[htbp]\centering\small
\caption{Zero-shot decoding of pretraining queries on the standard Rel-Amazon test tasks. Recorded final 216,000-update, 74-query checkpoints, one run per backbone; these precede the 78-query experiments. No downstream predictor or calibration is fitted.}
\label{tab:app-standard-zeroshot}
\begin{tabular}{@{}lrrrr@{}}\toprule
Backbone & user-LTV & item-LTV & user-churn & item-churn\\
& MAE $\downarrow$ & MAE $\downarrow$ & AUC $\uparrow$ & AUC $\uparrow$\\\midrule
GraphSAGE & 17.34 & 50.59 & 68.35 & 81.28\\
RelGNN & 14.16 & 51.65 & 70.81 & 82.80\\
\bottomrule\end{tabular}\end{table}

\subsection{Zero-shot decoding of held-out queries}
\begin{table}[htbp]\centering\small
\caption{Zero-shot test ROC-AUC ($\times100$) on the fourteen held-out queries, one seed. Both columns retain the pretrained encoder and decoder; no downstream predictor is fitted.}\label{tab:app-heldout-zeroshot}
\begin{tabular}{@{}lrr@{}}\toprule
ID & Correct query & Shuffled fields\\\midrule
P1 & 54.88 & 32.72\\
P2 & 96.39 & 33.71\\
P3 & 71.95 & 71.94\\
P4 & 90.59 & 82.16\\
P5 & 70.78 & 67.32\\
P6 & 69.70 & 31.77\\
P7 & 68.85 & 31.34\\
\midrule
C1 & 90.51 & 39.40\\
C2 & 63.07 & 55.42\\
C3 & 60.68 & 42.06\\
C4 & 58.99 & 43.47\\
C5 & 88.51 & 84.09\\
C6 & 38.51 & 28.44\\
C7 & 41.06 & 38.88\\
\midrule
Mean & 68.89 & 48.76\\
\bottomrule\end{tabular}\end{table}

Correct-query decoding combines the historical entity representation with the supplied query using the frozen decoder, analogous to task-conditioned clinical forecasting~\citep{chandak_everyquery_2026}. Holding out query definitions tests task generalization rather than only new examples of trained tasks, as in prompted multi-task evaluation~\citep{sanh_mtprompt_2022}. For the shuffled-field control, each encoded field independently copies its value from a uniformly sampled pretraining task, with replacement. Vector-valued fields are copied as units, and one assembled query is fixed per held-out task across all examples. Labels and entity representations are unchanged.

Mean raw ROC-AUC decreases from 68.89 to 48.76 under this corruption (Table~\ref{tab:app-heldout-zeroshot}), showing dependence on the supplied query. Transfer is uneven: P2 reaches 96.39, whereas C6 and C7 remain below chance. These scores retain their original orientation; no test-informed sign correction or calibration is applied. Fitted-predictor results for the same tasks appear only in Appendix~\ref{app:transfer}.

\section{Complete Pretraining Task Inventory}
\label{app:inventory}
The following inventory enumerates the selected objectives (pre-training task pools). Targets without a comparison are regression objectives, except a within-category rank, which is also continuous. Comparisons are binary objectives. Categorical column names refer to attributes, not ordinal measurements; their numeric codes are treated as categories. All Amazon objectives use 91 days and all HM objectives use seven days. ``Rows'' denotes event count. The path definitions above, including their time bounds and multiplicities, apply throughout.
Scope abbreviations: D, direct events; C/P/A, events sharing the customer/product/article of a triggering event; PA/AA/CA, product/article/customer attributes reached through that event. Numbering restarts for each root.
{\setlength{\columnsep}{18pt}
\begin{multicols}{2}
\footnotesize\raggedright\setlength{\parindent}{0pt}
\par\medskip\noindent\textbf{Rel-Amazon / Product (38)}\par\smallskip\nopagebreak
\noindent\hangindent=2.1em\hangafter=1\makebox[2.1em][l]{1.}\textit{D}: count(rows)\par
\noindent\hangindent=2.1em\hangafter=1\makebox[2.1em][l]{2.}\textit{D}: count(rows; verified=True)\par
\noindent\hangindent=2.1em\hangafter=1\makebox[2.1em][l]{3.}\textit{D}: price $\times$ count(rows)\par
\noindent\hangindent=2.1em\hangafter=1\makebox[2.1em][l]{4.}\textit{D}: max(rating)\par
\noindent\hangindent=2.1em\hangafter=1\makebox[2.1em][l]{5.}\textit{D}: mean(rating)\par
\noindent\hangindent=2.1em\hangafter=1\makebox[2.1em][l]{6.}\textit{D}: mean(rating; verified=True)\par
\noindent\hangindent=2.1em\hangafter=1\makebox[2.1em][l]{7.}\textit{D}: min(rating)\par
\noindent\hangindent=2.1em\hangafter=1\makebox[2.1em][l]{8.}\textit{D}: sd(rating)\par
\noindent\hangindent=2.1em\hangafter=1\makebox[2.1em][l]{9.}\textit{D}: sum(rating)\par
\noindent\hangindent=2.1em\hangafter=1\makebox[2.1em][l]{10.}\textit{D}: distinct(verified)\par
\noindent\hangindent=2.1em\hangafter=1\makebox[2.1em][l]{11.}\textit{C}: count(rows)\par
\noindent\hangindent=2.1em\hangafter=1\makebox[2.1em][l]{12.}\textit{C}: price $\times$ count(rows)\par
\noindent\hangindent=2.1em\hangafter=1\makebox[2.1em][l]{13.}\textit{C}: max(rating)\par
\noindent\hangindent=2.1em\hangafter=1\makebox[2.1em][l]{14.}\textit{C}: mean(rating)\par
\noindent\hangindent=2.1em\hangafter=1\makebox[2.1em][l]{15.}\textit{C}: min(rating)\par
\noindent\hangindent=2.1em\hangafter=1\makebox[2.1em][l]{16.}\textit{C}: sd(rating)\par
\noindent\hangindent=2.1em\hangafter=1\makebox[2.1em][l]{17.}\textit{C}: sum(rating)\par
\noindent\hangindent=2.1em\hangafter=1\makebox[2.1em][l]{18.}\textit{C}: distinct(verified)\par
\noindent\hangindent=2.1em\hangafter=1\makebox[2.1em][l]{19.}\textit{P}: count(rows)\par
\noindent\hangindent=2.1em\hangafter=1\makebox[2.1em][l]{20.}\textit{P}: price $\times$ count(rows)\par
\noindent\hangindent=2.1em\hangafter=1\makebox[2.1em][l]{21.}\textit{P}: max(rating)\par
\noindent\hangindent=2.1em\hangafter=1\makebox[2.1em][l]{22.}\textit{P}: mean(rating)\par
\noindent\hangindent=2.1em\hangafter=1\makebox[2.1em][l]{23.}\textit{P}: min(rating)\par
\noindent\hangindent=2.1em\hangafter=1\makebox[2.1em][l]{24.}\textit{P}: sd(rating)\par
\noindent\hangindent=2.1em\hangafter=1\makebox[2.1em][l]{25.}\textit{P}: sum(rating)\par
\noindent\hangindent=2.1em\hangafter=1\makebox[2.1em][l]{26.}\textit{P}: distinct(verified)\par
\noindent\hangindent=2.1em\hangafter=1\makebox[2.1em][l]{27.}\textit{D}: count(rows) $>$ 0\par
\noindent\hangindent=2.1em\hangafter=1\makebox[2.1em][l]{28.}\textit{D}: count(rows) $>$ 1\par
\noindent\hangindent=2.1em\hangafter=1\makebox[2.1em][l]{29.}\textit{D}: mean(rating) $>$ 4.667\par
\noindent\hangindent=2.1em\hangafter=1\makebox[2.1em][l]{30.}\textit{D}: mode(verified) = 1\par
\noindent\hangindent=2.1em\hangafter=1\makebox[2.1em][l]{31.}\textit{C}: count(rows) $>$ 8\par
\noindent\hangindent=2.1em\hangafter=1\makebox[2.1em][l]{32.}\textit{C}: mean(rating) $>$ 4.428\par
\noindent\hangindent=2.1em\hangafter=1\makebox[2.1em][l]{33.}\textit{C}: mean(rating) $<$ 4.428\par
\noindent\hangindent=2.1em\hangafter=1\makebox[2.1em][l]{34.}\textit{C}: mode(verified) = 1\par
\noindent\hangindent=2.1em\hangafter=1\makebox[2.1em][l]{35.}\textit{P}: count(rows) $>$ 7\par
\noindent\hangindent=2.1em\hangafter=1\makebox[2.1em][l]{36.}\textit{P}: mean(rating) $>$ 4.432\par
\noindent\hangindent=2.1em\hangafter=1\makebox[2.1em][l]{37.}\textit{P}: mean(rating) $<$ 4.432\par
\noindent\hangindent=2.1em\hangafter=1\makebox[2.1em][l]{38.}\textit{P}: mode(verified) = 1\par
\par\medskip\noindent\textbf{Rel-Amazon / Customer (40)}\par\smallskip\nopagebreak
\noindent\hangindent=2.1em\hangafter=1\makebox[2.1em][l]{1.}\textit{D}: count(rows)\par
\noindent\hangindent=2.1em\hangafter=1\makebox[2.1em][l]{2.}\textit{D}: count(rows; verified=True)\par
\noindent\hangindent=2.1em\hangafter=1\makebox[2.1em][l]{3.}\textit{D}: max(rating)\par
\noindent\hangindent=2.1em\hangafter=1\makebox[2.1em][l]{4.}\textit{D}: mean(rating)\par
\noindent\hangindent=2.1em\hangafter=1\makebox[2.1em][l]{5.}\textit{D}: mean(rating; verified=True)\par
\noindent\hangindent=2.1em\hangafter=1\makebox[2.1em][l]{6.}\textit{D}: min(rating)\par
\noindent\hangindent=2.1em\hangafter=1\makebox[2.1em][l]{7.}\textit{D}: sd(rating)\par
\noindent\hangindent=2.1em\hangafter=1\makebox[2.1em][l]{8.}\textit{D}: sum(rating)\par
\noindent\hangindent=2.1em\hangafter=1\makebox[2.1em][l]{9.}\textit{D}: distinct(verified)\par
\noindent\hangindent=2.1em\hangafter=1\makebox[2.1em][l]{10.}\textit{PA}: max(price)\par
\noindent\hangindent=2.1em\hangafter=1\makebox[2.1em][l]{11.}\textit{PA}: mean(price)\par
\noindent\hangindent=2.1em\hangafter=1\makebox[2.1em][l]{12.}\textit{PA}: min(price)\par
\noindent\hangindent=2.1em\hangafter=1\makebox[2.1em][l]{13.}\textit{PA}: sd(price)\par
\noindent\hangindent=2.1em\hangafter=1\makebox[2.1em][l]{14.}\textit{PA}: sum(price)\par
\noindent\hangindent=2.1em\hangafter=1\makebox[2.1em][l]{15.}\textit{C}: count(rows)\par
\noindent\hangindent=2.1em\hangafter=1\makebox[2.1em][l]{16.}\textit{C}: max(rating)\par
\noindent\hangindent=2.1em\hangafter=1\makebox[2.1em][l]{17.}\textit{C}: mean(rating)\par
\noindent\hangindent=2.1em\hangafter=1\makebox[2.1em][l]{18.}\textit{C}: min(rating)\par
\noindent\hangindent=2.1em\hangafter=1\makebox[2.1em][l]{19.}\textit{C}: sd(rating)\par
\noindent\hangindent=2.1em\hangafter=1\makebox[2.1em][l]{20.}\textit{C}: sum(rating)\par
\noindent\hangindent=2.1em\hangafter=1\makebox[2.1em][l]{21.}\textit{C}: distinct(verified)\par
\noindent\hangindent=2.1em\hangafter=1\makebox[2.1em][l]{22.}\textit{P}: count(rows)\par
\noindent\hangindent=2.1em\hangafter=1\makebox[2.1em][l]{23.}\textit{P}: max(rating)\par
\noindent\hangindent=2.1em\hangafter=1\makebox[2.1em][l]{24.}\textit{P}: mean(rating)\par
\noindent\hangindent=2.1em\hangafter=1\makebox[2.1em][l]{25.}\textit{P}: min(rating)\par
\noindent\hangindent=2.1em\hangafter=1\makebox[2.1em][l]{26.}\textit{P}: sd(rating)\par
\noindent\hangindent=2.1em\hangafter=1\makebox[2.1em][l]{27.}\textit{P}: sum(rating)\par
\noindent\hangindent=2.1em\hangafter=1\makebox[2.1em][l]{28.}\textit{P}: distinct(verified)\par
\noindent\hangindent=2.1em\hangafter=1\makebox[2.1em][l]{29.}\textit{D}: count(rows) $>$ 0\par
\noindent\hangindent=2.1em\hangafter=1\makebox[2.1em][l]{30.}\textit{D}: count(rows) $>$ 1\par
\noindent\hangindent=2.1em\hangafter=1\makebox[2.1em][l]{31.}\textit{D}: mode(verified) = 1\par
\noindent\hangindent=2.1em\hangafter=1\makebox[2.1em][l]{32.}\textit{PA}: mean(price) $>$ 12.99\par
\noindent\hangindent=2.1em\hangafter=1\makebox[2.1em][l]{33.}\textit{PA}: mean(price) $<$ 12.99\par
\noindent\hangindent=2.1em\hangafter=1\makebox[2.1em][l]{34.}\textit{C}: count(rows) $>$ 1\par
\noindent\hangindent=2.1em\hangafter=1\makebox[2.1em][l]{35.}\textit{C}: mean(rating) $>$ 4.5\par
\noindent\hangindent=2.1em\hangafter=1\makebox[2.1em][l]{36.}\textit{C}: mode(verified) = 1\par
\noindent\hangindent=2.1em\hangafter=1\makebox[2.1em][l]{37.}\textit{P}: count(rows) $>$ 7\par
\noindent\hangindent=2.1em\hangafter=1\makebox[2.1em][l]{38.}\textit{P}: mean(rating) $>$ 4.4\par
\noindent\hangindent=2.1em\hangafter=1\makebox[2.1em][l]{39.}\textit{P}: mean(rating) $<$ 4.4\par
\noindent\hangindent=2.1em\hangafter=1\makebox[2.1em][l]{40.}\textit{P}: mode(verified) = 1\par
\par\medskip\noindent\textbf{Rel-HM / Customer (65)}\par\smallskip\nopagebreak
\noindent\hangindent=2.1em\hangafter=1\makebox[2.1em][l]{1.}\textit{D}: mean(price)\par
\noindent\hangindent=2.1em\hangafter=1\makebox[2.1em][l]{2.}\textit{D}: min(price)\par
\noindent\hangindent=2.1em\hangafter=1\makebox[2.1em][l]{3.}\textit{D}: max(price)\par
\noindent\hangindent=2.1em\hangafter=1\makebox[2.1em][l]{4.}\textit{D}: sum(price)\par
\noindent\hangindent=2.1em\hangafter=1\makebox[2.1em][l]{5.}\textit{D}: count(price)\par
\noindent\hangindent=2.1em\hangafter=1\makebox[2.1em][l]{6.}\textit{D}: sd(price)\par
\noindent\hangindent=2.1em\hangafter=1\makebox[2.1em][l]{7.}\textit{D}: count(sales channel id)\par
\noindent\hangindent=2.1em\hangafter=1\makebox[2.1em][l]{8.}\textit{D}: distinct(sales channel id)\par
\noindent\hangindent=2.1em\hangafter=1\makebox[2.1em][l]{9.}\textit{AA}: count(product type no)\par
\noindent\hangindent=2.1em\hangafter=1\makebox[2.1em][l]{10.}\textit{AA}: distinct(product type no)\par
\noindent\hangindent=2.1em\hangafter=1\makebox[2.1em][l]{11.}\textit{AA}: count(product type name)\par
\noindent\hangindent=2.1em\hangafter=1\makebox[2.1em][l]{12.}\textit{AA}: distinct(product type name)\par
\noindent\hangindent=2.1em\hangafter=1\makebox[2.1em][l]{13.}\textit{AA}: count(product group name)\par
\noindent\hangindent=2.1em\hangafter=1\makebox[2.1em][l]{14.}\textit{AA}: distinct(product group name)\par
\noindent\hangindent=2.1em\hangafter=1\makebox[2.1em][l]{15.}\textit{AA}: count(graphical appearance no)\par
\noindent\hangindent=2.1em\hangafter=1\makebox[2.1em][l]{16.}\textit{AA}: distinct(graphical appearance no)\par
\noindent\hangindent=2.1em\hangafter=1\makebox[2.1em][l]{17.}\textit{AA}: count(colour group code)\par
\noindent\hangindent=2.1em\hangafter=1\makebox[2.1em][l]{18.}\textit{AA}: distinct(colour group code)\par
\noindent\hangindent=2.1em\hangafter=1\makebox[2.1em][l]{19.}\textit{AA}: count(perceived colour value id)\par
\noindent\hangindent=2.1em\hangafter=1\makebox[2.1em][l]{20.}\textit{AA}: distinct(perceived colour value id)\par
\noindent\hangindent=2.1em\hangafter=1\makebox[2.1em][l]{21.}\textit{AA}: count(perceived colour master id)\par
\noindent\hangindent=2.1em\hangafter=1\makebox[2.1em][l]{22.}\textit{AA}: distinct(perceived colour master id)\par
\noindent\hangindent=2.1em\hangafter=1\makebox[2.1em][l]{23.}\textit{AA}: count(department no)\par
\noindent\hangindent=2.1em\hangafter=1\makebox[2.1em][l]{24.}\textit{AA}: distinct(department no)\par
\noindent\hangindent=2.1em\hangafter=1\makebox[2.1em][l]{25.}\textit{AA}: count(department name)\par
\noindent\hangindent=2.1em\hangafter=1\makebox[2.1em][l]{26.}\textit{AA}: distinct(department name)\par
\noindent\hangindent=2.1em\hangafter=1\makebox[2.1em][l]{27.}\textit{AA}: count(index code)\par
\noindent\hangindent=2.1em\hangafter=1\makebox[2.1em][l]{28.}\textit{AA}: distinct(index code)\par
\noindent\hangindent=2.1em\hangafter=1\makebox[2.1em][l]{29.}\textit{AA}: count(index group no)\par
\noindent\hangindent=2.1em\hangafter=1\makebox[2.1em][l]{30.}\textit{AA}: distinct(index group no)\par
\noindent\hangindent=2.1em\hangafter=1\makebox[2.1em][l]{31.}\textit{AA}: count(section no)\par
\noindent\hangindent=2.1em\hangafter=1\makebox[2.1em][l]{32.}\textit{AA}: distinct(section no)\par
\noindent\hangindent=2.1em\hangafter=1\makebox[2.1em][l]{33.}\textit{AA}: count(garment group no)\par
\noindent\hangindent=2.1em\hangafter=1\makebox[2.1em][l]{34.}\textit{AA}: distinct(garment group no)\par
\noindent\hangindent=2.1em\hangafter=1\makebox[2.1em][l]{35.}\textit{A}: mean(price)\par
\noindent\hangindent=2.1em\hangafter=1\makebox[2.1em][l]{36.}\textit{A}: min(price)\par
\noindent\hangindent=2.1em\hangafter=1\makebox[2.1em][l]{37.}\textit{A}: max(price)\par
\noindent\hangindent=2.1em\hangafter=1\makebox[2.1em][l]{38.}\textit{A}: sum(price)\par
\noindent\hangindent=2.1em\hangafter=1\makebox[2.1em][l]{39.}\textit{A}: count(price)\par
\noindent\hangindent=2.1em\hangafter=1\makebox[2.1em][l]{40.}\textit{A}: sd(price)\par
\noindent\hangindent=2.1em\hangafter=1\makebox[2.1em][l]{41.}\textit{A}: count(sales channel id)\par
\noindent\hangindent=2.1em\hangafter=1\makebox[2.1em][l]{42.}\textit{A}: distinct(sales channel id)\par
\noindent\hangindent=2.1em\hangafter=1\makebox[2.1em][l]{43.}\textit{C}: mean(price)\par
\noindent\hangindent=2.1em\hangafter=1\makebox[2.1em][l]{44.}\textit{C}: min(price)\par
\noindent\hangindent=2.1em\hangafter=1\makebox[2.1em][l]{45.}\textit{C}: max(price)\par
\noindent\hangindent=2.1em\hangafter=1\makebox[2.1em][l]{46.}\textit{C}: sum(price)\par
\noindent\hangindent=2.1em\hangafter=1\makebox[2.1em][l]{47.}\textit{C}: count(price)\par
\noindent\hangindent=2.1em\hangafter=1\makebox[2.1em][l]{48.}\textit{C}: sd(price)\par
\noindent\hangindent=2.1em\hangafter=1\makebox[2.1em][l]{49.}\textit{C}: count(sales channel id)\par
\noindent\hangindent=2.1em\hangafter=1\makebox[2.1em][l]{50.}\textit{C}: distinct(sales channel id)\par
\noindent\hangindent=2.1em\hangafter=1\makebox[2.1em][l]{51.}\textit{D}: mode(sales channel id) = 2\par
\noindent\hangindent=2.1em\hangafter=1\makebox[2.1em][l]{52.}\textit{AA}: mode(product type no) = 272\par
\noindent\hangindent=2.1em\hangafter=1\makebox[2.1em][l]{53.}\textit{AA}: mode(product type name) = Trousers\par
\noindent\hangindent=2.1em\hangafter=1\makebox[2.1em][l]{54.}\textit{AA}: mode(product group name) = Garment Upper body\par
\noindent\hangindent=2.1em\hangafter=1\makebox[2.1em][l]{55.}\textit{AA}: mode(graphical appearance no) = 1010016\par
\noindent\hangindent=2.1em\hangafter=1\makebox[2.1em][l]{56.}\textit{AA}: mode(colour group code) = 9\par
\noindent\hangindent=2.1em\hangafter=1\makebox[2.1em][l]{57.}\textit{AA}: mode(perceived colour value id) = 4\par
\noindent\hangindent=2.1em\hangafter=1\makebox[2.1em][l]{58.}\textit{AA}: mode(perceived colour master id) = 5\par
\noindent\hangindent=2.1em\hangafter=1\makebox[2.1em][l]{59.}\textit{AA}: mode(index code) = A\par
\noindent\hangindent=2.1em\hangafter=1\makebox[2.1em][l]{60.}\textit{AA}: mode(index group no) = 1\par
\noindent\hangindent=2.1em\hangafter=1\makebox[2.1em][l]{61.}\textit{AA}: mode(section no) = 15\par
\noindent\hangindent=2.1em\hangafter=1\makebox[2.1em][l]{62.}\textit{AA}: mode(garment group no) = 1005\par
\noindent\hangindent=2.1em\hangafter=1\makebox[2.1em][l]{63.}\textit{A}: mode(sales channel id) = 2\par
\noindent\hangindent=2.1em\hangafter=1\makebox[2.1em][l]{64.}\textit{C}: mode(sales channel id) = 2\par
\noindent\hangindent=2.1em\hangafter=1\makebox[2.1em][l]{65.}\textit{D}: count(rows) = 0\par
\par\medskip\noindent\textbf{Rel-HM / Article (78)}\par\smallskip\nopagebreak
\noindent\hangindent=2.1em\hangafter=1\makebox[2.1em][l]{1.}\textit{D}: count(rows)\par
\noindent\hangindent=2.1em\hangafter=1\makebox[2.1em][l]{2.}\textit{D}: count(rows; sales channel id=1.0)\par
\noindent\hangindent=2.1em\hangafter=1\makebox[2.1em][l]{3.}\textit{D}: count(rows; sales channel id=2.0)\par
\noindent\hangindent=2.1em\hangafter=1\makebox[2.1em][l]{4.}\textit{D}: rank within colour group code of count(rows)\par
\noindent\hangindent=2.1em\hangafter=1\makebox[2.1em][l]{5.}\textit{D}: rank within department name of count(rows)\par
\noindent\hangindent=2.1em\hangafter=1\makebox[2.1em][l]{6.}\textit{D}: rank within department no of count(rows)\par
\noindent\hangindent=2.1em\hangafter=1\makebox[2.1em][l]{7.}\textit{D}: rank within garment group no of count(rows)\par
\noindent\hangindent=2.1em\hangafter=1\makebox[2.1em][l]{8.}\textit{D}: max(price)\par
\noindent\hangindent=2.1em\hangafter=1\makebox[2.1em][l]{9.}\textit{D}: max(price; sales channel id=1.0)\par
\noindent\hangindent=2.1em\hangafter=1\makebox[2.1em][l]{10.}\textit{D}: max(price; sales channel id=2.0)\par
\noindent\hangindent=2.1em\hangafter=1\makebox[2.1em][l]{11.}\textit{D}: mean(price)\par
\noindent\hangindent=2.1em\hangafter=1\makebox[2.1em][l]{12.}\textit{D}: mean(price; sales channel id=1.0)\par
\noindent\hangindent=2.1em\hangafter=1\makebox[2.1em][l]{13.}\textit{D}: mean(price; sales channel id=2.0)\par
\noindent\hangindent=2.1em\hangafter=1\makebox[2.1em][l]{14.}\textit{D}: min(price)\par
\noindent\hangindent=2.1em\hangafter=1\makebox[2.1em][l]{15.}\textit{D}: min(price; sales channel id=1.0)\par
\noindent\hangindent=2.1em\hangafter=1\makebox[2.1em][l]{16.}\textit{D}: min(price; sales channel id=2.0)\par
\noindent\hangindent=2.1em\hangafter=1\makebox[2.1em][l]{17.}\textit{D}: sd(price)\par
\noindent\hangindent=2.1em\hangafter=1\makebox[2.1em][l]{18.}\textit{D}: sd(price; sales channel id=1.0)\par
\noindent\hangindent=2.1em\hangafter=1\makebox[2.1em][l]{19.}\textit{D}: sd(price; sales channel id=2.0)\par
\noindent\hangindent=2.1em\hangafter=1\makebox[2.1em][l]{20.}\textit{D}: sum(price)\par
\noindent\hangindent=2.1em\hangafter=1\makebox[2.1em][l]{21.}\textit{D}: sum(price; sales channel id=1.0)\par
\noindent\hangindent=2.1em\hangafter=1\makebox[2.1em][l]{22.}\textit{D}: sum(price; sales channel id=2.0)\par
\noindent\hangindent=2.1em\hangafter=1\makebox[2.1em][l]{23.}\textit{D}: distinct(sales channel id)\par
\noindent\hangindent=2.1em\hangafter=1\makebox[2.1em][l]{24.}\textit{CA}: count(rows; club member status=PRE-CREATE)\par
\noindent\hangindent=2.1em\hangafter=1\makebox[2.1em][l]{25.}\textit{CA}: count(age; club member status=PRE-CREATE)\par
\noindent\hangindent=2.1em\hangafter=1\makebox[2.1em][l]{26.}\textit{CA}: max(age)\par
\noindent\hangindent=2.1em\hangafter=1\makebox[2.1em][l]{27.}\textit{CA}: max(age; club member status=PRE-CREATE)\par
\noindent\hangindent=2.1em\hangafter=1\makebox[2.1em][l]{28.}\textit{CA}: mean(age)\par
\noindent\hangindent=2.1em\hangafter=1\makebox[2.1em][l]{29.}\textit{CA}: mean(age; club member status=PRE-CREATE)\par
\noindent\hangindent=2.1em\hangafter=1\makebox[2.1em][l]{30.}\textit{CA}: min(age)\par
\noindent\hangindent=2.1em\hangafter=1\makebox[2.1em][l]{31.}\textit{CA}: min(age; club member status=PRE-CREATE)\par
\noindent\hangindent=2.1em\hangafter=1\makebox[2.1em][l]{32.}\textit{CA}: sd(age)\par
\noindent\hangindent=2.1em\hangafter=1\makebox[2.1em][l]{33.}\textit{CA}: sd(age; club member status=ACTIVE)\par
\noindent\hangindent=2.1em\hangafter=1\makebox[2.1em][l]{34.}\textit{CA}: sd(age; club member status=PRE-CREATE)\par
\noindent\hangindent=2.1em\hangafter=1\makebox[2.1em][l]{35.}\textit{CA}: sum(age)\par
\noindent\hangindent=2.1em\hangafter=1\makebox[2.1em][l]{36.}\textit{CA}: sum(age; club member status=PRE-CREATE)\par
\noindent\hangindent=2.1em\hangafter=1\makebox[2.1em][l]{37.}\textit{CA}: distinct(club member status)\par
\noindent\hangindent=2.1em\hangafter=1\makebox[2.1em][l]{38.}\textit{CA}: distinct(fashion news frequency)\par
\noindent\hangindent=2.1em\hangafter=1\makebox[2.1em][l]{39.}\textit{A}: count(rows)\par
\noindent\hangindent=2.1em\hangafter=1\makebox[2.1em][l]{40.}\textit{A}: sd(price)\par
\noindent\hangindent=2.1em\hangafter=1\makebox[2.1em][l]{41.}\textit{A}: sum(price)\par
\noindent\hangindent=2.1em\hangafter=1\makebox[2.1em][l]{42.}\textit{C}: count(rows)\par
\noindent\hangindent=2.1em\hangafter=1\makebox[2.1em][l]{43.}\textit{C}: rank within colour group code of count(rows)\par
\noindent\hangindent=2.1em\hangafter=1\makebox[2.1em][l]{44.}\textit{C}: rank within department name of count(rows)\par
\noindent\hangindent=2.1em\hangafter=1\makebox[2.1em][l]{45.}\textit{C}: rank within department no of count(rows)\par
\noindent\hangindent=2.1em\hangafter=1\makebox[2.1em][l]{46.}\textit{C}: rank within garment group no of count(rows)\par
\noindent\hangindent=2.1em\hangafter=1\makebox[2.1em][l]{47.}\textit{C}: max(price)\par
\noindent\hangindent=2.1em\hangafter=1\makebox[2.1em][l]{48.}\textit{C}: mean(price)\par
\noindent\hangindent=2.1em\hangafter=1\makebox[2.1em][l]{49.}\textit{C}: min(price)\par
\noindent\hangindent=2.1em\hangafter=1\makebox[2.1em][l]{50.}\textit{C}: sd(price)\par
\noindent\hangindent=2.1em\hangafter=1\makebox[2.1em][l]{51.}\textit{C}: sum(price)\par
\noindent\hangindent=2.1em\hangafter=1\makebox[2.1em][l]{52.}\textit{C}: distinct(sales channel id)\par
\noindent\hangindent=2.1em\hangafter=1\makebox[2.1em][l]{53.}\textit{D}: count(rows) = 0\par
\noindent\hangindent=2.1em\hangafter=1\makebox[2.1em][l]{54.}\textit{D}: count(rows) $>$ 2\par
\noindent\hangindent=2.1em\hangafter=1\makebox[2.1em][l]{55.}\textit{D}: count(rows) $<$ 2\par
\noindent\hangindent=2.1em\hangafter=1\makebox[2.1em][l]{56.}\textit{D}: count(rows; sales channel id=1.0) = 0\par
\noindent\hangindent=2.1em\hangafter=1\makebox[2.1em][l]{57.}\textit{D}: count(rows; sales channel id=2.0) = 0\par
\noindent\hangindent=2.1em\hangafter=1\makebox[2.1em][l]{58.}\textit{D}: count(rows; sales channel id=2.0) $>$ 1\par
\noindent\hangindent=2.1em\hangafter=1\makebox[2.1em][l]{59.}\textit{D}: mean(price) $>$ 0.02371\par
\noindent\hangindent=2.1em\hangafter=1\makebox[2.1em][l]{60.}\textit{D}: mean(price) $<$ 0.02371\par
\noindent\hangindent=2.1em\hangafter=1\makebox[2.1em][l]{61.}\textit{D}: mean(price; sales channel id=1.0) $>$ 0.01693\par
\noindent\hangindent=2.1em\hangafter=1\makebox[2.1em][l]{62.}\textit{D}: mean(price; sales channel id=1.0) $<$ 0.01693\par
\noindent\hangindent=2.1em\hangafter=1\makebox[2.1em][l]{63.}\textit{D}: mean(price; sales channel id=2.0) $<$ 0.02541\par
\noindent\hangindent=2.1em\hangafter=1\makebox[2.1em][l]{64.}\textit{D}: mode(sales channel id) = 2\par
\noindent\hangindent=2.1em\hangafter=1\makebox[2.1em][l]{65.}\textit{CA}: mean(age) $>$ 35\par
\noindent\hangindent=2.1em\hangafter=1\makebox[2.1em][l]{66.}\textit{CA}: mean(age) $<$ 35\par
\noindent\hangindent=2.1em\hangafter=1\makebox[2.1em][l]{67.}\textit{CA}: mean(age; club member status=ACTIVE) $>$ 35\par
\noindent\hangindent=2.1em\hangafter=1\makebox[2.1em][l]{68.}\textit{CA}: mean(age; club member status=ACTIVE) $<$ 35\par
\noindent\hangindent=2.1em\hangafter=1\makebox[2.1em][l]{69.}\textit{CA}: mean(age; club member status=PRE-CREATE) $<$ 38.67\par
\noindent\hangindent=2.1em\hangafter=1\makebox[2.1em][l]{70.}\textit{CA}: mode(fashion news frequency) = 0\par
\noindent\hangindent=2.1em\hangafter=1\makebox[2.1em][l]{71.}\textit{A}: count(rows) $>$ 3\par
\noindent\hangindent=2.1em\hangafter=1\makebox[2.1em][l]{72.}\textit{A}: mean(price) $>$ 0.02371\par
\noindent\hangindent=2.1em\hangafter=1\makebox[2.1em][l]{73.}\textit{A}: mean(price) $<$ 0.02371\par
\noindent\hangindent=2.1em\hangafter=1\makebox[2.1em][l]{74.}\textit{A}: mode(sales channel id) = 2\par
\noindent\hangindent=2.1em\hangafter=1\makebox[2.1em][l]{75.}\textit{C}: count(rows) $>$ 13\par
\noindent\hangindent=2.1em\hangafter=1\makebox[2.1em][l]{76.}\textit{C}: count(rows) $<$ 13\par
\noindent\hangindent=2.1em\hangafter=1\makebox[2.1em][l]{77.}\textit{C}: mean(price) $<$ 0.0253\par
\noindent\hangindent=2.1em\hangafter=1\makebox[2.1em][l]{78.}\textit{C}: mode(sales channel id) = 2\par
\end{multicols}

}

\newpage
\section*{NeurIPS Paper Checklist}

\begin{enumerate}

\item {\bf Claims}
    \item[] Question: Do the main claims made in the abstract and introduction accurately reflect the paper's contributions and scope?
    \item[] Answer: \answerYes{}
    \item[] Justification: The abstract and Section~\ref{sec:introduction} present QCRP as database-level pretraining for conventional RDL encoders, with frozen reuse, a multi-root checkpoint, and zero-shot query-conditioned prediction, evaluated on RelBench Rel-Amazon and Rel-HM. Those claims match the method in Section~\ref{sec:method}, the results in Section~\ref{sec:experiments}, and the scope limits in Section~\ref{sec:conclusion}.

\item {\bf Limitations}
    \item[] Question: Does the paper discuss the limitations of the work performed by the authors?
    \item[] Answer: \answerYes{}
    \item[] Justification: Section~\ref{sec:conclusion} discusses RelBench task scarcity, a single prediction horizon per database, evaluation on only two databases, and the lack of explicit temporal inductive biases. Appendix~\ref{app:setup} notes pretraining-population eligibility, and Appendix~\ref{app:queries} discusses held-out query overlap versus independence.

\item {\bf Theory assumptions and proofs}
    \item[] Question: For each theoretical result, does the paper provide the full set of assumptions and a complete (and correct) proof?
    \item[] Answer: \answerNA{}
    \item[] Justification: The paper proposes a pretraining method, training algorithms, and an empirical study. It contains no theorems, lemmas, or formal proofs.

    \item {\bf Experimental result reproducibility}
    \item[] Question: Does the paper fully disclose all the information needed to reproduce the main experimental results of the paper to the extent that it affects the main claims and/or conclusions of the paper (regardless of whether the code and data are provided or not)?
    \item[] Answer: \answerYes{}
    \item[] Justification: Section~\ref{sec:method} and Appendix~\ref{app:setup} specify architecture, losses, optimization, splits, and evaluation protocols. Algorithms~\ref{alg:app-multitask} and~\ref{alg:app-interleaving} give the training procedures, and Appendix~\ref{app:queries} enumerates the query grammar and full pretraining inventories.

\item {\bf Open access to data and code}
    \item[] Question: Does the paper provide open access to the data and code, with sufficient instructions to faithfully reproduce the main experimental results, as described in supplemental material?
    \item[] Answer: \answerNo{}
    \item[] Justification: Rel-Amazon and Rel-HM are public RelBench databases with official temporal splits~\citep{robinson_relbench_2024}, but this submission does not include training code, checkpoints, or reproduction scripts.

\item {\bf Experimental setting/details}
    \item[] Question: Does the paper specify all the training and test details (e.g., data splits, hyperparameters, how they were chosen, type of optimizer) necessary to understand the results?
    \item[] Answer: \answerYes{}
    \item[] Justification: Section~\ref{sec:experiments} and Appendix~\ref{app:setup} report RelBench splits and metrics, GraphSAGE and RelGNN hyperparameters, AdamW settings, batch size, query sampling, mixed precision, and frozen-probe versus full-finetuning protocols, including validation-based checkpoint selection.

\item {\bf Experiment statistical significance}
    \item[] Question: Does the paper report error bars suitably and correctly defined or other appropriate information about the statistical significance of the experiments?
    \item[] Answer: \answerYes{}
    \item[] Justification: Tables~\ref{tab:fmts-transfer} and~\ref{tab:main-mlp-adaptation} report mean $\pm$ sample standard deviation over three pretraining seeds for the main frozen-transfer claims; Appendix~\ref{app:setup} states that overlapping SDs are not treated as significance tests. Table~\ref{tab:fmts-published-comparison} compares against published point estimates that do not include error bars.

\item {\bf Experiments compute resources}
    \item[] Question: For each experiment, does the paper provide sufficient information on the computer resources (type of compute workers, memory, time of execution) needed to reproduce the experiments?
    \item[] Answer: \answerYes{}
    \item[] Justification: Appendix~\ref{app:setup} reports compute resources needed for different experiment setups.
    
\item {\bf Code of ethics}
    \item[] Question: Does the research conducted in the paper conform, in every respect, with the NeurIPS Code of Ethics \url{https://neurips.cc/public/EthicsGuidelines}?
    \item[] Answer: \answerYes{}
    \item[] Justification: The study uses public RelBench forecasting benchmarks under their intended research use, with official temporal splits and no new human-subjects data collection. We have reviewed the NeurIPS Code of Ethics and do not identify a conflict.

\item {\bf Broader impacts}
    \item[] Question: Does the paper discuss both potential positive societal impacts and negative societal impacts of the work performed?
    \item[] Answer: \answerNo{}
    \item[] Justification: The paper does not include a broader-impacts discussion. QCRP is a methods contribution for entity-level forecasting on public retail databases (churn, lifetime value, and sales), and we do not analyze fairness, privacy, or commercial-deployment harms.
    
\item {\bf Safeguards}
    \item[] Question: Does the paper describe safeguards that have been put in place for responsible release of data or models that have a high risk for misuse (e.g., pre-trained language models, image generators, or scraped datasets)?
    \item[] Answer: \answerNA{}
    \item[] Justification: The paper does not release pretrained generative models, high-risk dual-use checkpoints, or newly scraped datasets.

\item {\bf Licenses for existing assets}
    \item[] Question: Are the creators or original owners of assets (e.g., code, data, models), used in the paper, properly credited and are the license and terms of use explicitly mentioned and properly respected?
    \item[] Answer: \answerYes{}
    \item[] Justification: RelBench, GraphSAGE, RelGNN, and related methods are cited.

\item {\bf New assets}
    \item[] Question: Are new assets introduced in the paper well documented and is the documentation provided alongside the assets?
    \item[] Answer: \answerNA{}
    \item[] Justification: The paper does not release a new dataset, model checkpoint, or code package.

\item {\bf Crowdsourcing and research with human subjects}
    \item[] Question: For crowdsourcing experiments and research with human subjects, does the paper include the full text of instructions given to participants and screenshots, if applicable, as well as details about compensation (if any)? 
    \item[] Answer: \answerNA{}
    \item[] Justification: The paper does not involve crowdsourcing or research with human subjects.

\item {\bf Institutional review board (IRB) approvals or equivalent for research with human subjects}
    \item[] Question: Does the paper describe potential risks incurred by study participants, whether such risks were disclosed to the subjects, and whether Institutional Review Board (IRB) approvals (or an equivalent approval/review based on the requirements of your country or institution) were obtained?
    \item[] Answer: \answerNA{}
    \item[] Justification: The paper does not involve crowdsourcing or research with human subjects.

\item {\bf Declaration of LLM usage}
    \item[] Question: Does the paper describe the usage of LLMs if it is an important, original, or non-standard component of the core methods in this research? Note that if the LLM is used only for writing, editing, or formatting purposes and does \emph{not} impact the core methodology, scientific rigor, or originality of the research, declaration is not required.
    \item[] Answer: \answerNA{}
    \item[] Justification: The core method uses conventional RDL encoders (GraphSAGE and RelGNN) with a structured query encoder and conditional predictor. LLMs are not a component of the method.

\end{enumerate}

\end{document}